\documentclass[conference]{IEEEtran}
\ifCLASSINFOpdf
\else
\fi
\usepackage{graphicx}
\usepackage{listings}
\usepackage{xcolor}
\usepackage{siunitx}
\usepackage{booktabs}
\usepackage{multirow}
\usepackage{algorithm}
\usepackage{algorithmic}
\usepackage{subfigure}
\usepackage{threeparttable}
\graphicspath{{fig/}}
\begin{document}
	%
	\title{CNNLab: a Novel Parallel Framework for Neural Networks using GPU and FPGA
		\\{\LARGE --- a Practical Study with Trade-off Analysis}
	}
	\author{\IEEEauthorblockN{Maohua Zhu, Liu Liu}
		\IEEEauthorblockA{Electrical and Computer Engineering, UCSB\\
		Email:\{maohuazhu,liu\_liu\}@umail.ucsb.edu}
		\and
		\IEEEauthorblockN{Chao Wang}
		\IEEEauthorblockA{Computer Sciecne,USTC\\
		Email:cswang@ustc.edu.cn	}
		\and
		\IEEEauthorblockN{Yuan Xie}
		\IEEEauthorblockA{Electrical and Computer Engineering,UCSB\\
			Email:yuanxie@ece.ucsb.edu}}
	\maketitle
	
	\begin{abstract}
		Designing and implementing efficient, provably correct parallel neural network processing is challenging. Existing high-level parallel abstractions like MapReduce are insufficiently expressive while low-level tools like MPI and Pthreads leave ML experts repeatedly solving the same design challenges. However, the diversity and large-scale data size have posed a significant challenge to construct a flexible and high-performance implementation of deep learning neural networks. To improve the performance and maintain the scalability, we present CNNLab, a novel deep learning framework using GPU and FPGA-based accelerators. CNNLab provides a uniform programming model to users so that the hardware implementation and the scheduling are invisible to the programmers. At runtime, CNNLab leverages the trade-offs between GPU and FPGA before offloading the tasks to the accelerators. Experimental results on the state-of-the-art Nvidia K40 GPU and Altera DE5 FPGA board demonstrate that the CNNLab can provide a universal framework with efficient support for diverse applications without increasing the burden of the programmers. Moreover, we analyze the detailed quantitative performance, throughput, power, energy, and performance density for both approaches. Experimental results leverage the trade-offs between GPU and FPGA and provide useful practical experiences for the deep learning research community.
	\end{abstract}
	

	%
	\IEEEpeerreviewmaketitle

	\section{Introduction}
	
	In the past several years, machine learning has become pervasive in various research fields and commercial applications with achieved satisfactory products. In particular, the emerging of deep learning speeded up the development of machine learning and artificial intelligence. Consequently, deep learning has become a research hotspot in research organizations and the companies\cite{deeplearning}. In general, deep learning uses a multi-layer neural network model to extract high-level features into a combination of low-level abstractions to find the distributed data features, to solve complex problems in machine learning. Currently, the most widely used neural models of deep learning are Deep Neural Networks (DNNs) and Convolution Neural Networks (CNNs), which have excellent capability in solving picture recognition, voice recognition, and other complex machine learning tasks.
	
	However, with the increasing accuracy requirements and complexity for the practical applications, the size of the networks becomes explosively large scale (for example, the Google cat-recognizing system has 1 Billion neuronal connections). The explosive volume of data makes the data centers quite power consuming. Therefore, it poses significant challenges to implementing high-performance deep learning networks with low power cost, especially for large-scale deep learning neural network models.
	
	The state-of-the-art means for accelerating deep learning algorithms are Field-Programmable Gate Array (FPGA) \cite{dnnfpga15}, Application Specific Integrated Circuit (ASIC) \cite{Diannao}, and Graphic Processing Unit (GPU) \cite{micro15gpu}. GPU has been well recognized for its high performance in massive computing capacity. Compared with GPU acceleration, hardware accelerators like FPGA and ASIC can achieve at least satisfying performance with lower power consumption. However, both FPGA and ASIC have relatively limited computing resources, memory, and I/O bandwidths, therefore it is challenging to develop complex and massive deep neural networks using hardware accelerators. Up to now, the problem of providing efficient middleware support for different architectures has not been adequately solved.
	
	Another challenge is the diversity and programming in deep learning applications. Due to the design complexity of the deep learning algorithms, architectures, and accelerators, it requires significant programming effort to make satisfying utilization of the accelerations in diverse application domains. If the computation is solved by the programmer manually, the quality of scheduling depends on the experiences of the programmer, who has limited knowledge of the hardware. To alleviate the burden of the high-level programmers, we can demonstrate the effectiveness of an efficient middleware support in deep learning research paradigm. To tackle these problems, in this paper, we present CNNLab, which is a middleware architecture targeting the state-of-the-art GPU and FPGA-based deep learning acceleration engines. Our main contributions are the following:
	
	\begin{itemize}
		\item We introduce a novel framework into the state-of-the-art architecture for deep learning applications. Applications can be mapped heterogeneous accelerators within a well-structured interface to improve the flexibility and scalability. CNNLab maps the applications into computing kernels using CUDA and OpenCL programming interfaces.
		\item CNNLab is based on a heterogeneous hybrid system which includes the software processor, GPU accelerator, and FPGA-based hardware accelerator to speed up the kernel computational parts of deep learning algorithms. In particular, we utilize an efficient middleware support to bridge the gap between high-level neural networks and hardware accelerators.
		\item We construct a hardware prototype using state-of-the-art Nvidia K40 GPU and Altera FPGA platforms. Experimental results on real hardware demonstrate that CNNLab can achieve remarkable speedup with insignificant overheads. More importantly, to explore the trade-offs between different implementations using FPGA and GPU, we leverage the trade-offs by analyzing the quantitative results for the running time, throughput, power, energy and performance density of the accelerators.
	\end{itemize}
	
	The rest of the paper is organized as follows: Section II summarizes the problem description and motivation. Section III discusses the CNNLab architecture, including architecture, the programming model, and the hierarchical layers. After that, Section IV describes the GPU and FPGA implementation of the accelerators. We analyze the detailed results and discusses the trade-offs between the GPU and FPGA based implementations. Section V outlines the related study of the neural network accelerators. Finally, Section VI describes the conclusion and further works.

	\begin{figure}[!t]
		\centering
		\includegraphics[width=3in]{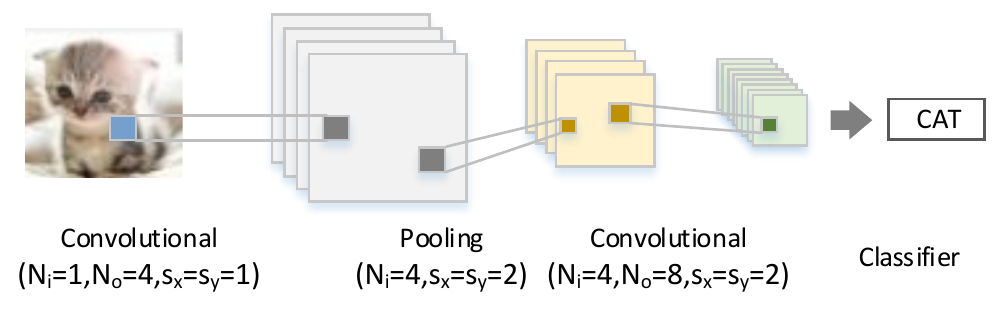}
		\caption{Neural network hierarchy containing convolutional, pooling, and classifier layers.}
		\label{motivation}
	\end{figure}

	\section{Problem Description and Motivation}
	Deep Learning has recently gained great popularity in the machine learning community due to their potential in solving previously difficult learning problems. Even though Deep and Convolutional Neural Networks have diverse forms, they share similar properties that a generic description can be formalized. First, these algorithms consist of a large number of layers, which are normally executed in sequence so they can be implemented and evaluated separately. Second, each layer usually contains several sub-layers called feature maps; we then use the terms feature input maps and feature output maps. Overall, there are three main kinds of layers: most of the hierarchy is composed of convolutional and pooling layers, and there is a classifier at the top of the network consisting of one or multiple layers. The role of convolutional layers is to apply one or several local filters to data from the input layer. Consequently, the connectivity between the input and output feature map is local. Consider the case where the input is an image, and the convolution is a 2D transform between a $Kx × Ky$ subset of the input layer and a kernel of the same dimensions, as illustrated in Fig. \ref{motivation}. The kernel values are the synaptic weights between an input layer and an output layer. In general, a DNN has two computational steps, including prediction process and training. Prediction process is a feedforward computation which computes the output for each given input with network settings. Training process includes pre-training which locally tunes the connection weights between the units in adjacent layers and global training which globally tunes the connection weights with the back-propagation (BP) algorithm.
	
	Firstly, we introduce the prediction process which is a feedforward computation. The process computes in accordance with traditional neural network layer by layer, and the outputs of current layer are the inputs of the next layer. The deep neural network is composed of an input layer, an output layer, and multiple hidden layers to get representations of the data with multiple levels of abstraction. The prediction computation of DNNs is a bottom-up feed-forward process, where the output of the lower layer is the input of its upper layer. To present the prediction computation of deep neural networks clearly, we consider a layer (called L) with $N_o$ neurons, the lower layer of which has $N_i$ neurons. The connectivity between layers is full, so each pair of neurons own their private weight values, resulting an $N_i$x$N_o$ weight matrix. L reads in $N_i$ inputs ($x_1, x_2, ..., x_{N_i}$) from its lower layer, and then produces $N_o$ outputs ($y_1, y_2, ..., y_{N_o}$). The calculation of a neuron $y_k$ (k = 1, 2, ...$N_o$) in L can be represented as $f(\sum_{j=1}^{N_i}W_\mathit{jk}x_\mathit{j} + b_k)$ where $x_j$ is a neuron of the lower layer, \emph f is the activation function, \emph W$_\mathit{jk}$ is the weight coefficient of \emph x$_\mathit{j}$ and \emph y$_\mathit{k}$, and \emph b$_\mathit{k}$ means the offset value. Since \(\sum_{j=1}^{N_i}W_\mathit{jk}x_\mathit{j} \) can be regarded as a multiplication between a row vector $\overrightarrow{X}$ and a column vector $\overrightarrow{W_\mathit{k}}$, the computation of the whole layer can be formalized as a vector-matrix multiplication and activation function process, shown as Equation $\left(\ref{equ:matrix}\right)$:
	
	\begin{equation}\label{equ:matrix}\mathit{ \overrightarrow{Y} = f(\overrightarrow{X} * W) }\end{equation}
	
	where:
	\begin{equation}\ \mathit{\overrightarrow{X} = (1, x_1, x_2, ..., x_{N_i}), ~~ \overrightarrow{Y} = (y_1, y_2, ..., y_{N_o})}\end{equation}
	\begin{equation}\ \mathit W =
	\left(
	\begin{array}{cccc}
	b_\mathit{1} & b_\mathit{2} & \cdots\ & b_\mathit{N_o}\\
	W_\mathit{11} & W_\mathit{12} & \cdots\ & W_\mathit{1{N_o}}\\
	\vdots\ &  \vdots\ & \ddots\ &  \vdots\\\
	W_\mathit{{N_i}1} & W_\mathit{{N_i}2} & \cdots\ & W_\mathit{{N_i}{N_o}}\\
	\end{array}
	\right)
	\end{equation}
	\begin{equation}\mathit{ f(x) = \frac {1}{1+e^{-x}}},~~\mbox {sigmoid~function}\end{equation}
	
	To accelerate the kernel function of the CNN processing, this paper presents CNNLab architecture, which uses both GPU and FPGA as the platform to explore the tradeoff of the performance/power metrics.

	\section{The CNNLab Abstraction}
	\begin{figure}[!t]
		\centering
		\includegraphics[width=3in]{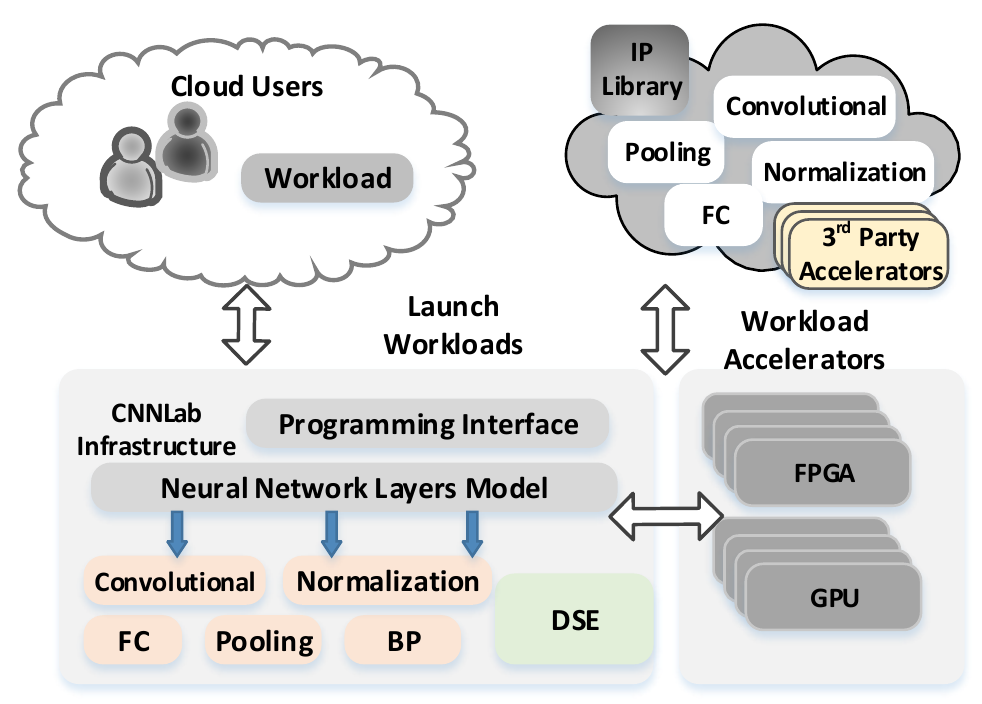}
		\caption{High level CNNLab architecture with corresponding neural network layers model. The NN processing are decomposed into layers and scheduled either on the GPU or FPGA-based accelerators.}
		\label{framework}
	\end{figure}
	
	\subsection{Data Model and Processing Flow}
	In this section, we present the modeling framework of the CNNLab architecture. Fig. \ref{framework} illustrates the common infrastructure of CNNLab with high-level perspective. The front-end cloud users access the CNNlab platform via a uniform programming interface with the user definitions of the CNN model, whereby each layer is packaged and exposed in an API-like manner. Meanwhile, in the back-end, each layer is composed of specific functionalities provided by software libraries and resource pool through consistent communication interfaces. The functionality for each layer is combined with multiple data resources with a sequence of input/output parameters. It should be noted that the detailed composition and the middleware scheduling is invisible to front-end software users. At runtime, the application is first decomposed into multiple layers under the definition of specific parameters, which are then scheduled at runtime. Whenever a pending layer has obtained its requisite input parameters, it can be offloaded to a particular accelerator for immediate execution.

\begin{figure}[!t]
		\centering
		\includegraphics[width=3in]{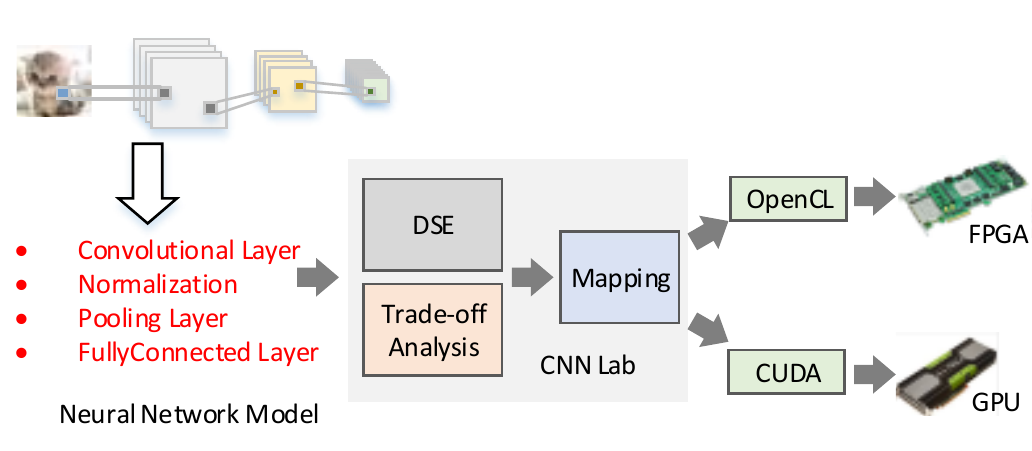}
		\caption{General Processing Flow of the CNNLab.}
		\label{flow}
	\end{figure}

	Fig. \ref{flow} presents a high-level overview of CNNLab processing flow in following steps. As the first step, the Deep Learning Specialist provides as inputs a high-level description of a ConvNet architecture together with information about the target CNNLab platform with the aid of a general layer-based model that we designed. Next, the structure of the NN input model will undergo the design space exploration and trade-off analysis in the middleware support, considering the requirements of the application. The design space is searched, and this process yields a succession of hardware mappings of the NN model onto the particular FPGA-based or GPU-based platforms, using OpenCL or CUDA programming interface, respectively.
	
	\subsection{User Defined Computation}
	To describe the CNN model, we define each layer associated with a tuple of parameters. Currently, the following types of layers are supported, which are the ones that have been most commonly used in the ConvNet literature:
	
	\subsubsection{\textbf{Convolutional Layer}}
	The model of Convolutional Layer is abstracted as
	\begin{equation}
	<M_{I},M_{K},M_{O},S,T>
	\end{equation}

	where
	\begin{itemize}
		\item $M_{I}$ and $M_{O}$ are the input/output matrix of each convolutional layer, which includes height $\times$ width $\times$ dimension.
		\item $M_{K}$ refers to the kernel size each accelerator can processed with, which includes height $\times$ width $\times$ dimension.
		\item $S$ is the stride which defines the step between successive convolution windows.
		\item $T$ is the type of nonlinear function to be applied, e.g. sigmoid, tanh or ReLU.
	\end{itemize}
	

	\subsubsection{\textbf{Normalization Layer}}
	The model of Normalization Layer is abstracted as
	\begin{equation}
	< M_{I},T,S,\alpha,\beta >
	\end{equation}
	
	where
	\begin{itemize}
		\item $M_{I}$ is the input matrix of the normalization layer, which includes height $\times$ width $\times$ dimension.
		\item $T$ is the type of normalization operation to be applied.
		\item $S$ is the local size applied in the nonlinear layer.
		\item $\alpha$ and $\beta$ are the parameters used in LRN computation.
	\end{itemize}

	\subsubsection{\textbf{Pooling Layer}}
	
	The model of Pooling Layer is abstracted as
	\begin{equation}
	<M_{I},M_{O},T,S,N>
	\end{equation}
	
	where
	\begin{itemize}
		\item $M_{I}$ and $M_{O}$ are the input/output matrix of the pooling layer, which includes height $\times$ width $\times$ dimension.
		\item T is the type of pooling operation to be applied, i.e. either max or average.
		\item N is the number of pooling kernels in the pooling layer.
		\item S is the stride which defines the step between successive pooling windows.
	\end{itemize}
	
	\begin{figure}[!t]
		\centering
		\includegraphics[width=3in]{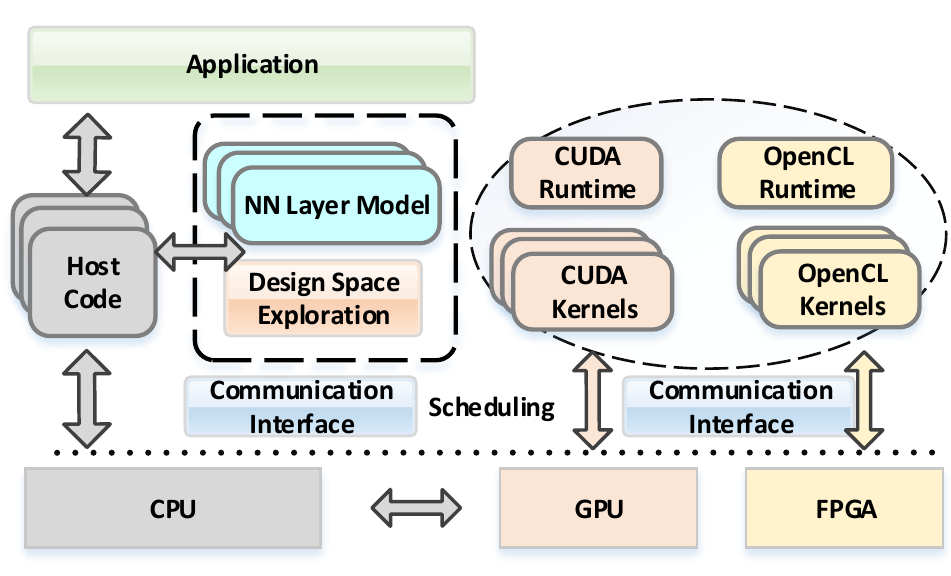}
		\caption{General Programming Framework}
		\label{programming}
	\end{figure}
	
	\begin{figure*}[!t]
		\centering
		\includegraphics[width=7in]{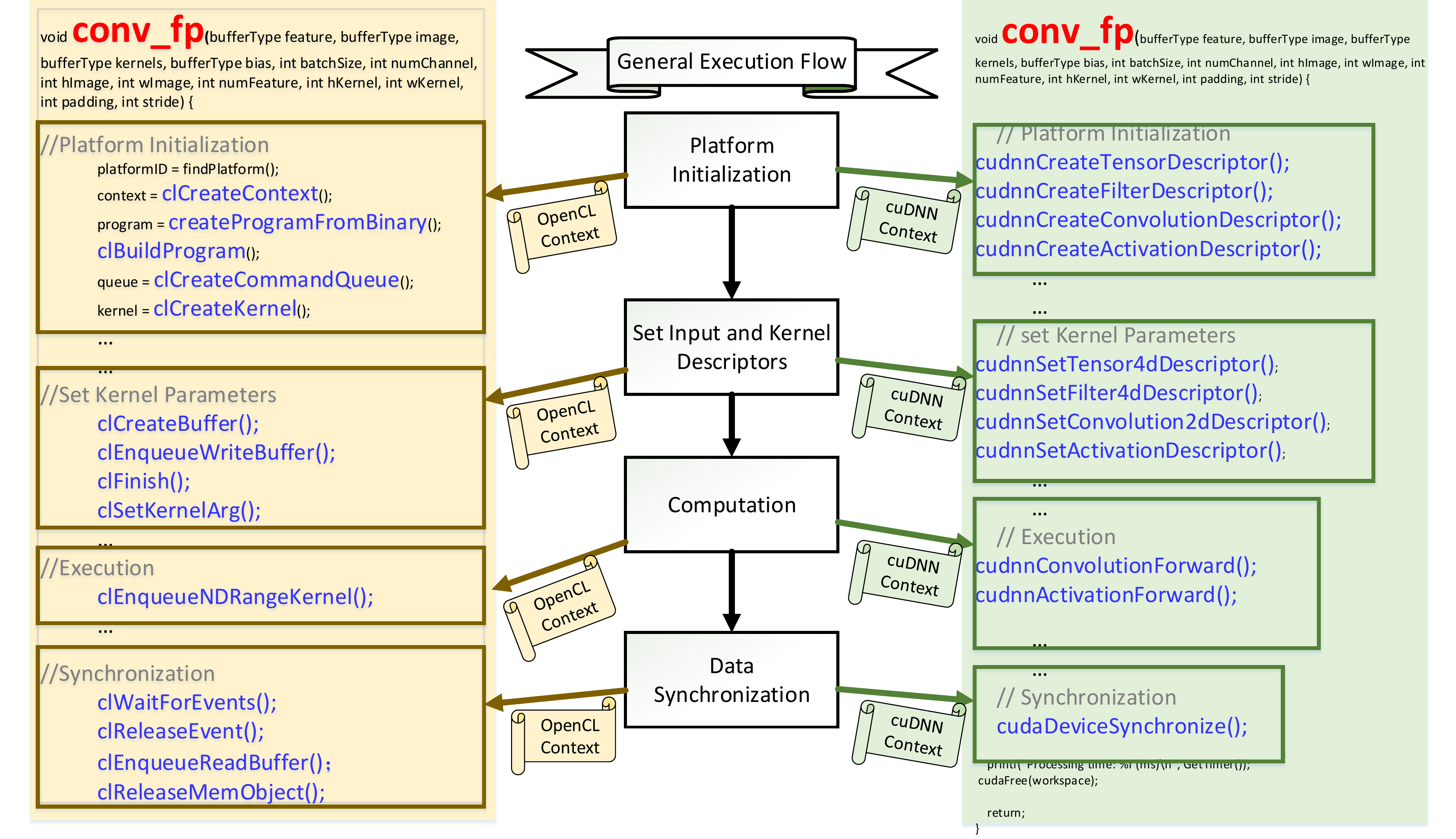}
		\caption{An Example Code Snippet for Convolutional Layer using cuDNN and OpenCL kernels}
		\label{snippet}
	\end{figure*}
	
	\subsubsection{\textbf{FC Layer}}
	The model of FC Layer is abstracted as
	\begin{equation}
	<M_{I},K_{O}>
	\end{equation}
	
	where
	\begin{itemize}
		\item $M_{I}$ is the input matrix of each convolutional layer, which includes height $\times$ width $\times$ dimension for FC-dropout layer.
		\item $K_{O}$ stands for the output of the FC layer.
	\end{itemize}

	\subsection{Programming Model and Wrapper}

	Based on the flexible programming framework in CNN lab, the tasks can be distributed to either GPU and FPGA-based accelerators. In particular, the middleware support should provide a uniform runtime for different hardware architectures. Fig. \ref{programming} illustrates the general programming framework based on the runtime kernels. The API forwards the requests via the scheduling middleware on the host code. The host code can offload part of the execution threads to CUDA kernels or OpenCL kernels, depending on the accelerator architecture and the design space exploration. Different kernels share a virtual memory space for communication among the parameters of accelerators. The scheduling process and run-time support are invisible to the programmers as the API provides an efficient bridge for high-level applications to hardware implementations. In particular, we use two example code snippets using both CUDA (cuDNN V5) and OpenCL as a demonstration.
	
	Fig. \ref{snippet} illustrates an example code segment using GPU and FPGA-based accelerators. The general processing flow includes following steps: 1) Platform Initialization, 2) Set Input and Kernel Descriptors, 3) Computation using Accelerators, and 4) Data Synchronization after Execution. To offload the kernels to the GPU and FPGA-based accelerators, the OpenCL contexts and cuDNN contexts are invoked with specific primitives. The code segments is general so as to be ported to other GPU or FPGA based hardware platforms.
	
	\begin{table*}[!t]
		\renewcommand{\arraystretch}{1.3}
		\caption{Description of the Experimental Neural Network Model }
		\label{model}
		\centering
		\begin{tabular}{|c||c||c|}
			\hline
			\textbf{Layer Name}    &\textbf{Layer Type}&    \textbf{Description} \\ \hline
			Conv1 &    Conv-ReLU&            Input: 3x224x224, Kernel: 96x3x11x11, Output: 96x55x55, Stride: 4 \\ \hline
			Conv2 &    Conv-ReLU    &            Input: 96x27x27, Kernel: 256x96x5x5, Output: 256x27x27, Stride: 1\\ \hline
			Conv3&    Conv-ReLU    &            Input: 256x13x13, Kernel: 384x256x3x3, Output: 384x13x13, Stride: 1\\ \hline
			Conv4&    Conv-ReLU    &                Input: 384x13x13, Kernel: 384x384x3x3, Output: 384x13x13, Stride: 1 \\ \hline
			Conv5&    Conv-ReLU    &                Input: 384x13x13, Kernel: 256x384x3x3, Output: 256x13x13, Stride: 1 \\ \hline
			FC6&    FC-dropout    &            Input: 256x6x6, Output: 4096 \\ \hline
			FC7&    FC-dropout    &            Input: 4096, Output: 4096\\ \hline
			FC8&    FC-softmax    &            Input: 4096, Output: 1000\\ \hline
		\end{tabular}
		
	\end{table*}
	
	\begin{figure*}
		\begin{minipage}{0.32\linewidth}
			\centerline{\includegraphics[width=6cm]{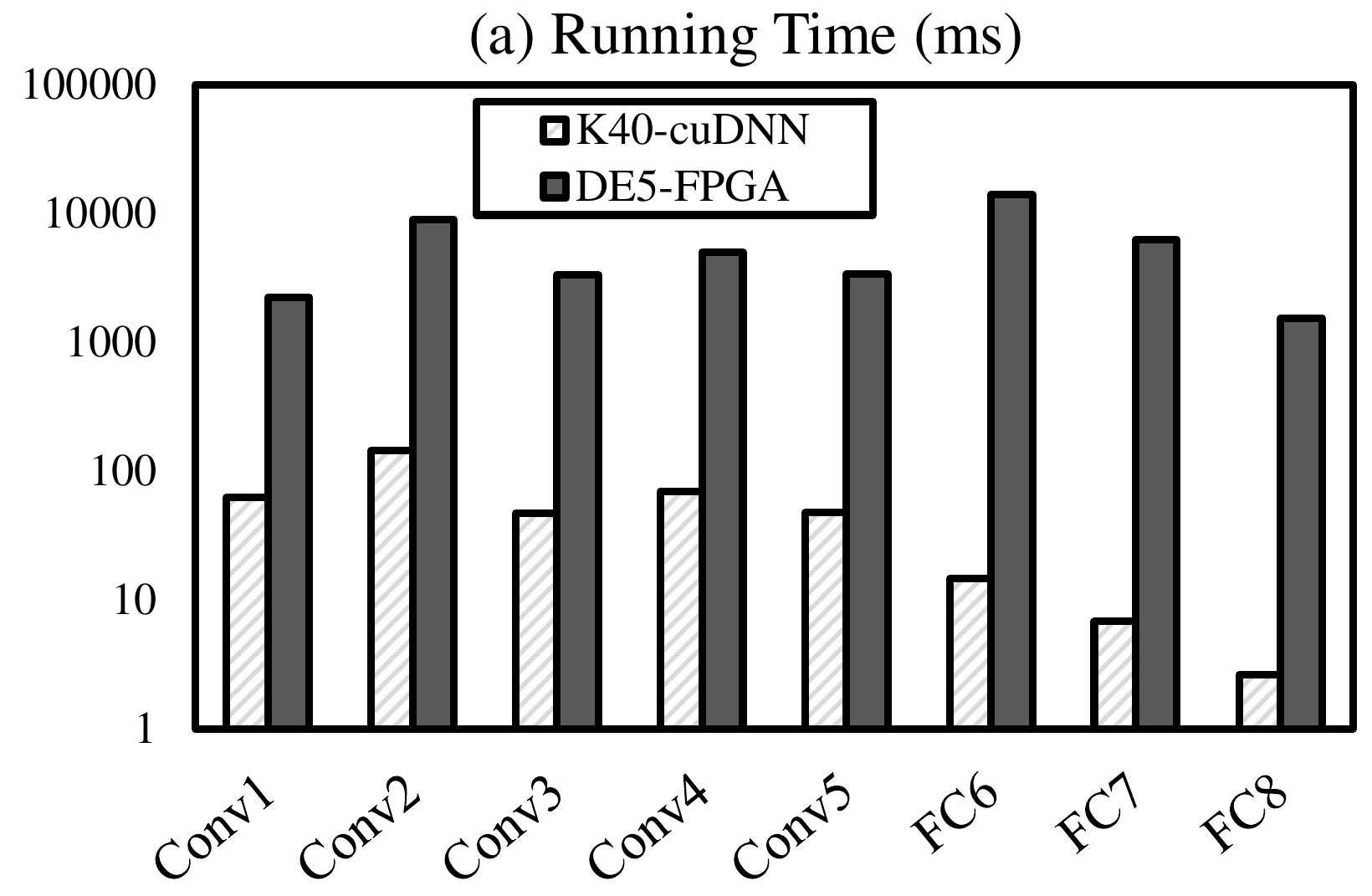}}
			\centerline
			\centerline{\includegraphics[width=6cm]{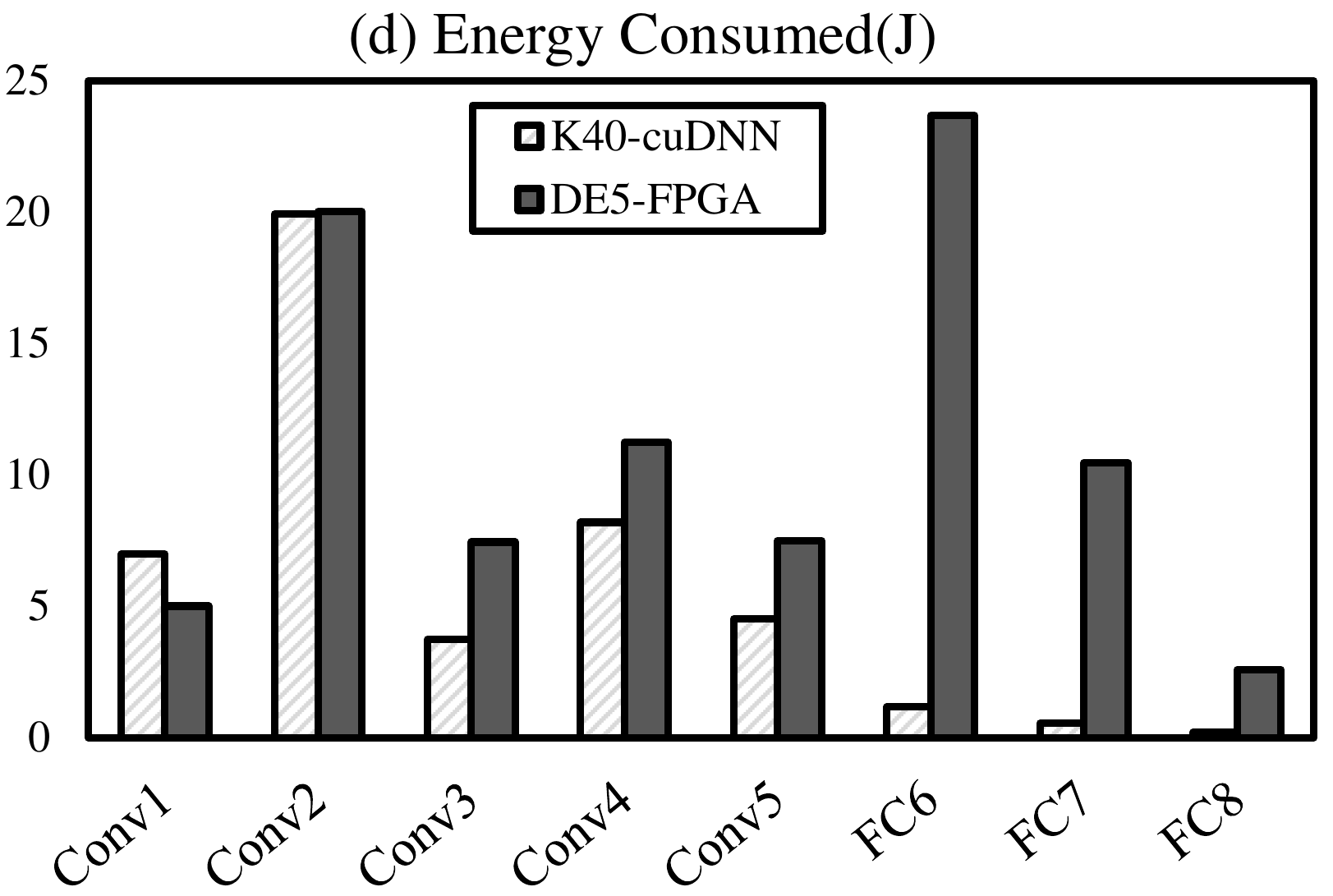}}
			
		\end{minipage}
		\hfill
		\begin{minipage}{0.32\linewidth}
			\centerline{\includegraphics[width=6cm]{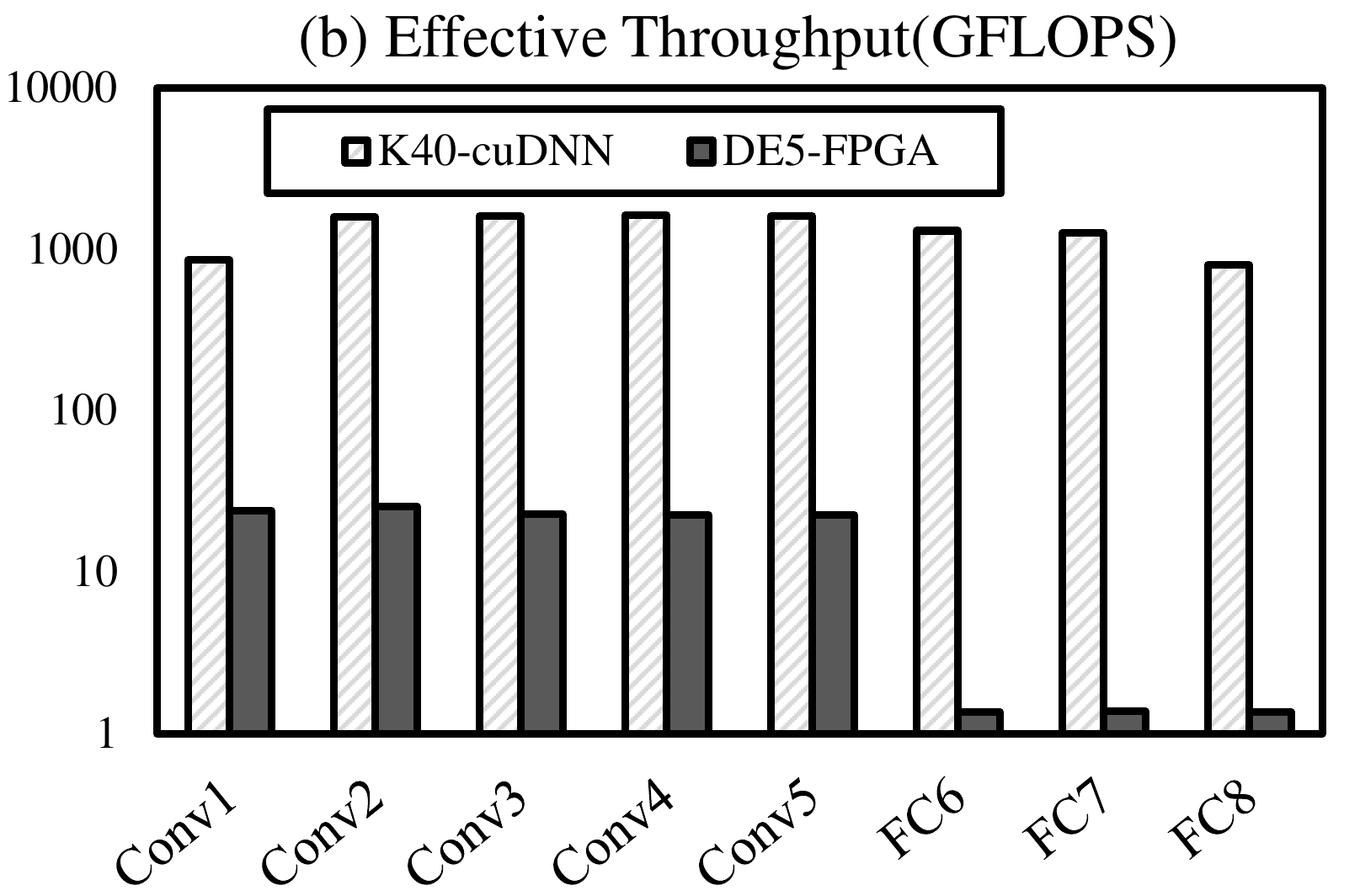}}
			\centerline
			\centerline{\includegraphics[width=6cm]{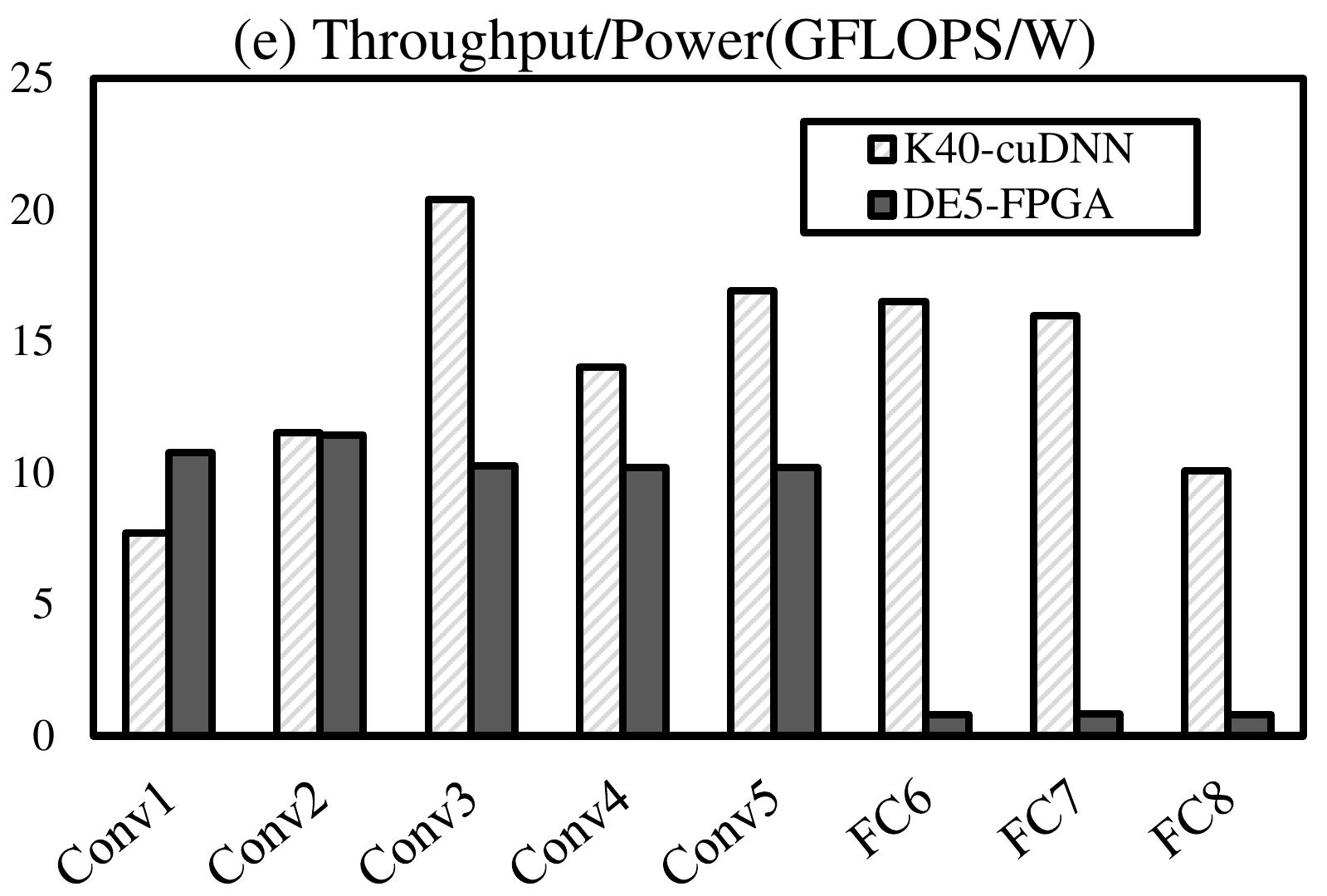}}
			
		\end{minipage}
		\hfill
		\begin{minipage}{0.32\linewidth}
			\centerline{\includegraphics[width=6cm]{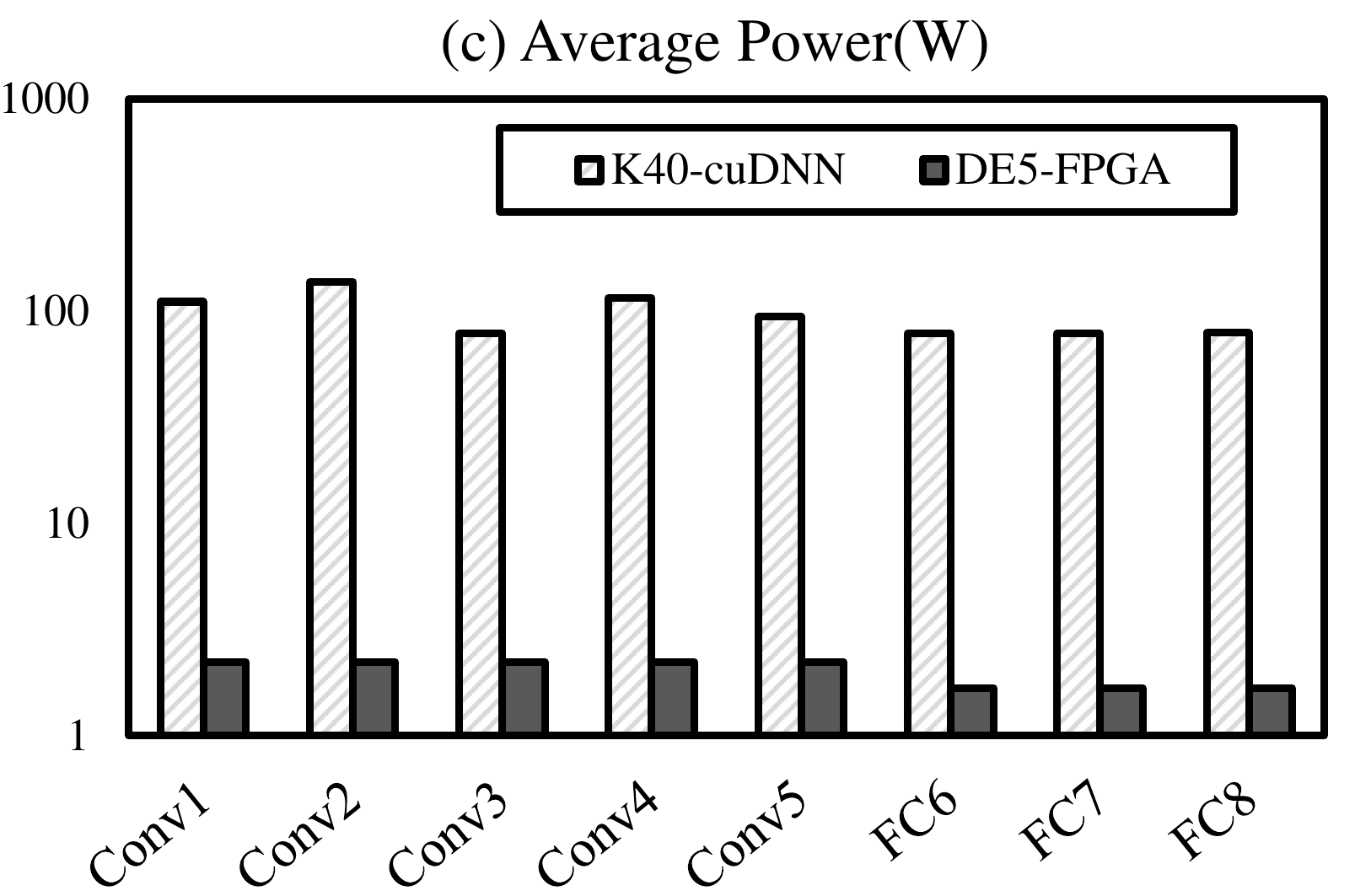}}
			\centerline
			\centerline{\includegraphics[width=6cm]{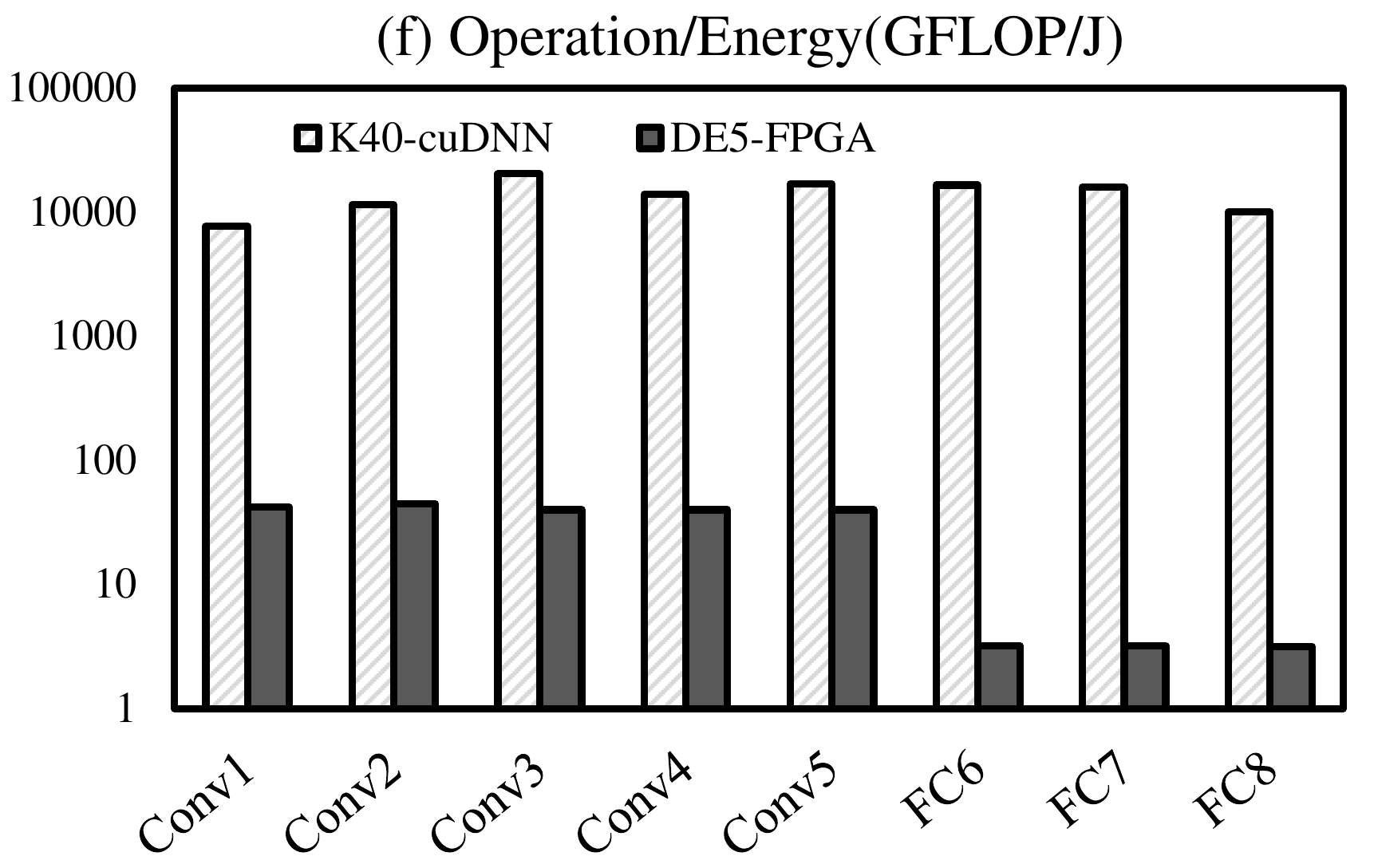}}
			
		\end{minipage}
		\hfill
		\caption{Evaluation and Trade-off Analysis between GPU and FPGA based Acceleration Engines.}
		\label{comparison}
	\end{figure*}
	
	\section{Experiments Results and Trade-off Analysis}
	
	\subsection{Platform Setup and Network Description}
	
	To measure the performance and overheads of CNNLab architecture, we implemented a hardware prototype with hybrid heterogeneous systems:
	
	\textbf{CPU: } An Intel Corei7-4770 processor clocked at 3.4 GHz was used as the CPU controller. The CPU processor assigns the computational tasks to the acceleration kernels via PCIE X8 edge connector for interconnection.
	
	\textbf{FPGA: } An Intel-Altera DE5 was used to implement the design of deep learning module. Altera Quartus II toolchain to evaluate the speedup and hardware cost, as well as the PowerPlay to estimate the power consumption. The running frequencies of the accelerators on FPGA range from 171MHz to 304MHz (see Table \ref{resource} for detail).
	
	\textbf{GPU: }A Nvidia K40 was used as the GPU accelerator, with 12,288 MB memory capacity, peak bandwidth of device memory at 288 GB/s, and peak single-precision floating point performance at 4.29TFLOPS. We use the state-of-the-art cuDNN V5 as the CUDA programming models (released in April 2016).

	Table \ref{model} introduces the experimental neural network model, including 5 Convolutional Layers and 3 FC Layers. We use ReLU as the nonlinear function in the Convolutional layer. For each layer, the parameters introduced in Section III.B is realized with different configurations.

	\begin{table*}[!t]
		\renewcommand{\arraystretch}{1.3}
		\caption{Network Description of GPU Models }
		\label{gpupara}
		\centering
		\begin{tabular}{|c||c||c||c||c||c|}
			\hline
			\textbf{Process} & \textbf{Layer Name}    &\textbf{Layer Type}&\textbf{fp operations per image} &    \textbf{Device}&        \textbf{Description} \\ \hline
			\multirow{6}{*}{Forward}
			&    FC6&    FC-dropout    &        75497472&    K40-cudnn    &    Input: 256x6x6, Output: 4096    \\  \cline{2-6}
			&    FC7&    FC-dropout    &        33554432&    K40-cudnn    &    Input: 4096, Output: 4096        \\ \cline{2-6}
			&    FC8&    FC-softmax    &        8192000    &    K40-cudnn    &    Input: 4096, Output: 1000         \\ \cline{2-6}
			&     FC6&    FC-dropout    &        75497472&    K40-cublas    &    Input: 256x6x6, Output: 4096    \\ \cline{2-6}
			&    FC7&    FC-dropout    &        33554432&    K40-cublas    &    Input: 4096, Output: 4096        \\ \cline{2-6}
			&    FC8&    FC-softmax    &        8192000    &    K40-cublas    &    Input: 4096, Output: 1000        \\ \hline
			\multirow{6}{*}{Backward}
			&    FC6&    FC-dropout    &    150994944    &K40-cudnn        &Input: 256x6x6, Output: 4096     \\  \cline{2-6}
			&    FC7&    FC-dropout    &    67108864    &K40-cudnn        &Input: 4096, Output: 4096        \\  \cline{2-6}
			&    FC8&    FC-softmax    &    16384000    &K40-cudnn        &Input: 4096, Output: 1000        \\ \cline{2-6}
			&    FC6&    FC-dropout    &    150994944    &K40-cublas            &Input: 256x6x6, Output: 4096    \\  \cline{2-6}
			&    FC7&    FC-dropout    &    67108864    &K40-cublas            &Input: 4096, Output: 4096        \\  \cline{2-6}
			&    FC8&    FC-softmax    &    16384000    &K40-cublas            &Input: 4096, Output: 1000        \\ \hline
		\end{tabular}
	\end{table*}

	\subsection{Results Analysis and Trade-offs between FPGA and GPU}
	
	We analyze the trade-offs between the two approaches on the following aspects: Performance(including execution time and throughput), Cost (including power and energy), and performance density (including throughput per watt, and throughput per joule) respectively.
	
	\textbf{Performance.} Fig. \ref{comparison} (a) presents the running time for the eight layers. GPU has better performance than FPGA on all the layers, and the speedup can achieve up to 1000x for FC layers. Regarding the eight layers, the speedup for convolutional layers (1-5) is lower than the FC layers (6-8), which contains matrix multiplication operations. We also evaluate and compare the throughput, as illustrated in (b). Results are similar to the running time that the GPU can achieve significant higher throughput than FPGA. For example, the peak throughput for GPU is 1632 GFLOPS in Conv 4 layer, while the peak throughput for FPGA is only 25.56 GFLOPS in Conv 2 layer.
	
	\textbf{Power and Energy.} To establish the cost model for both approaches, we evaluate the power and energy consumption for GPU and FPGA-based accelerators. Fig. \ref{comparison} (c) illustrates the comparison of the power cost. The average power for GPU is 97W while the power of FPGA-based accelerator for the convolutional layer is only 2.23W. To this end, FPGA is power saving due to the limited hardware resources and low working frequency (300MHz). Concerning the energy, both approaches have similar energy consumption when running convolutional layers. For example, the average energy for FPGA is 10.24J, while GPU cost 8.67J on average. In comparison, FPGA takes significantly higher energy for FC layers than GPU, as presented in (d). For FC layers, the average energy consumption for FPGA is 12.24J, while GPU only takes 0.64J on average. Results demonstrate that GPU can achieve better energy efficiency on FC layers due to the optimization of matrix multiplication operations.
	
	\textbf{Performance Density.} Based on the performance and the power cost, we derive the performance density for both methods. First, for Throughput/Power metrics, GPU and FPGA has similar performance density in convolutional layers, that the GPU achieves 14.12 GFLOPS/W while FPGA gets 10.58 GFLOPS/W. For the FC layers, GPU substantially outperforms FPGA by achieving the average density at 14.20 GFLOPS/W, while FPGA only has 0.82 GFLOPS/W. Regarding the energy metrics, we measure the Operation/Energy (GFLOP/J) as the metric. In this case, GPU far outperforms the GPU by achieving 14732 GFLOP/J for all the layers on average, while FPGA only gets 41.35 GFLOP/J for the convolutional layer on average, and 3.19 GFLOP/J for FC layers.
	
	Above results reveals that GPU can achieve higher speedup and throughput while FPGA saves more power consumption. Regarding the energy consumption and performance density, both approaches get similar results for convolutional computation, and GPU outperforms FPGA in the calculation for FC layers.

	\begin{figure}
		\begin{minipage}{0.47\linewidth}
			\centerline{\includegraphics[width=4.5cm]{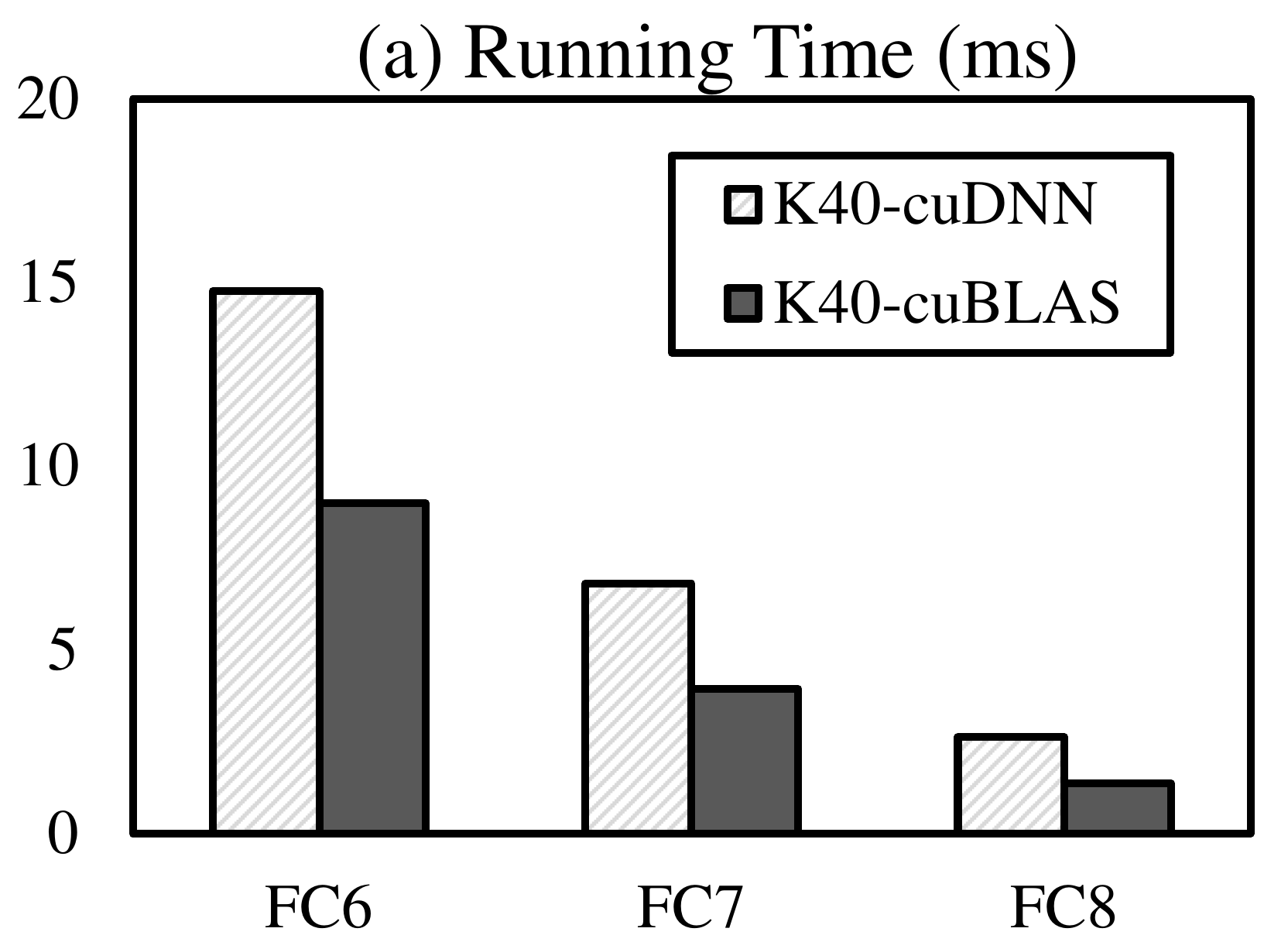}}
			\centerline{}
			\centerline{\includegraphics[width=4.5cm]{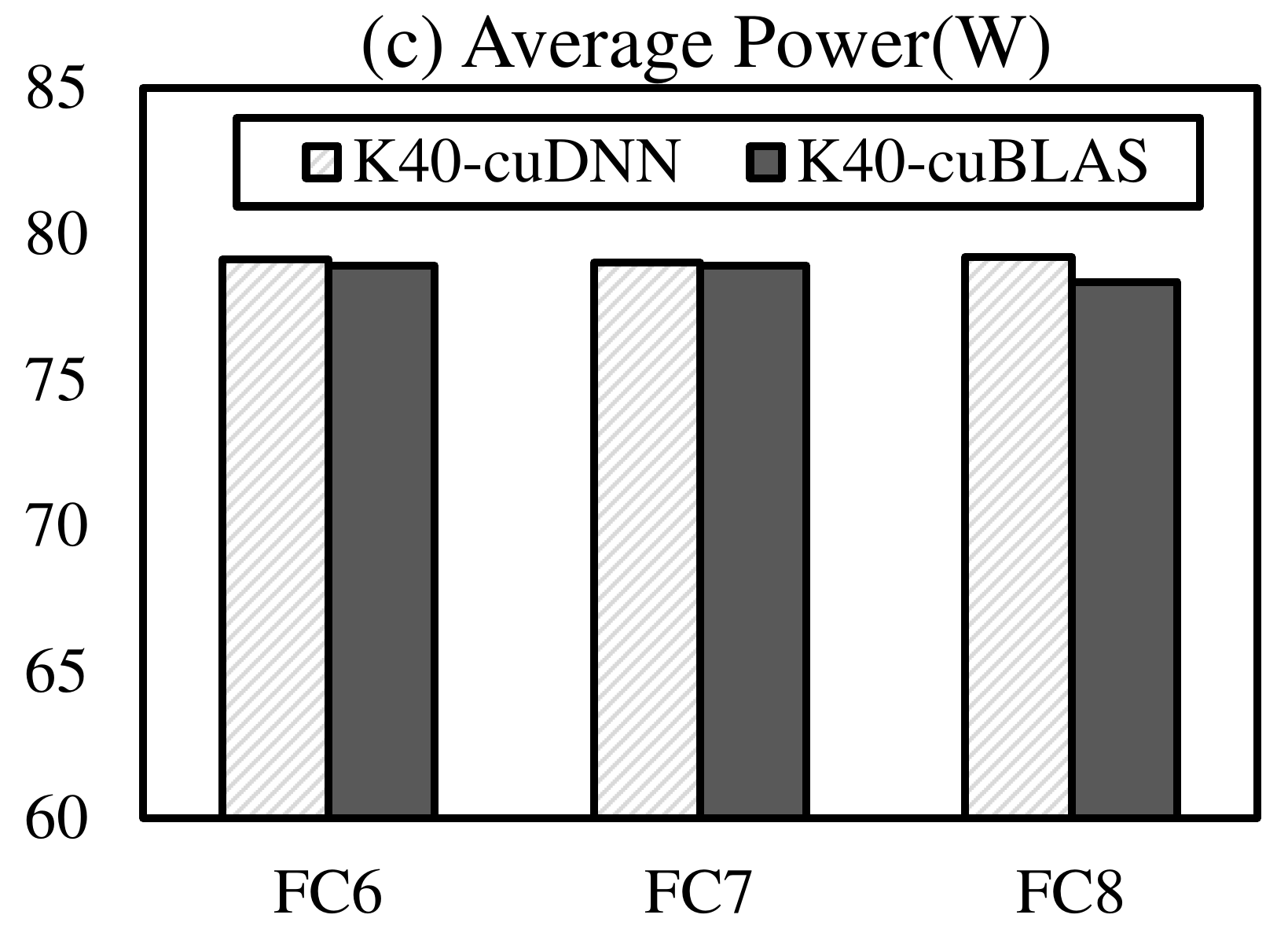}}
			\centerline{}
			\centerline{\includegraphics[width=4.5cm]{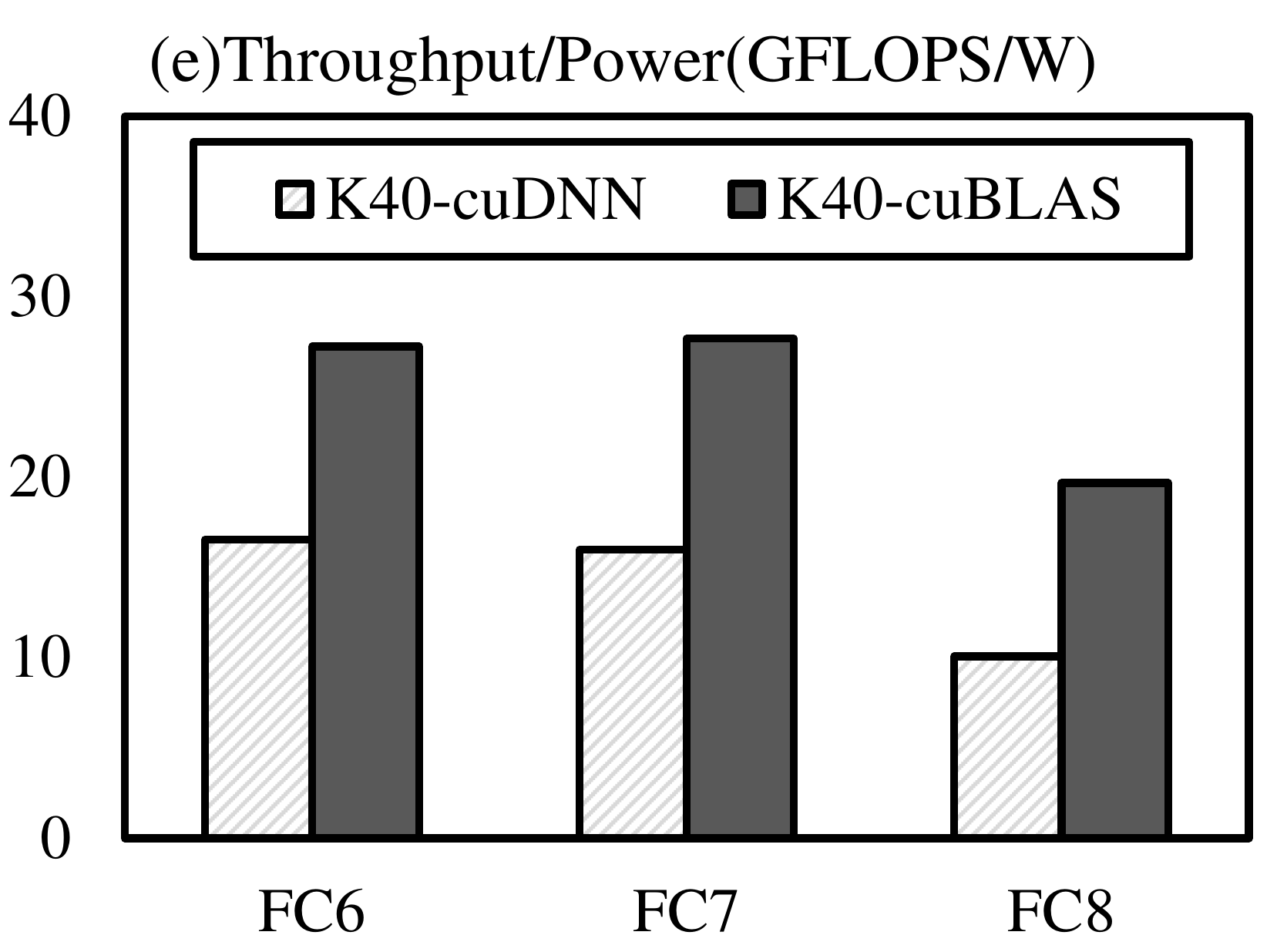}}
		\end{minipage}
		\hfill
		\begin{minipage}{0.47\linewidth}
			\centerline{\includegraphics[width=4.5cm]{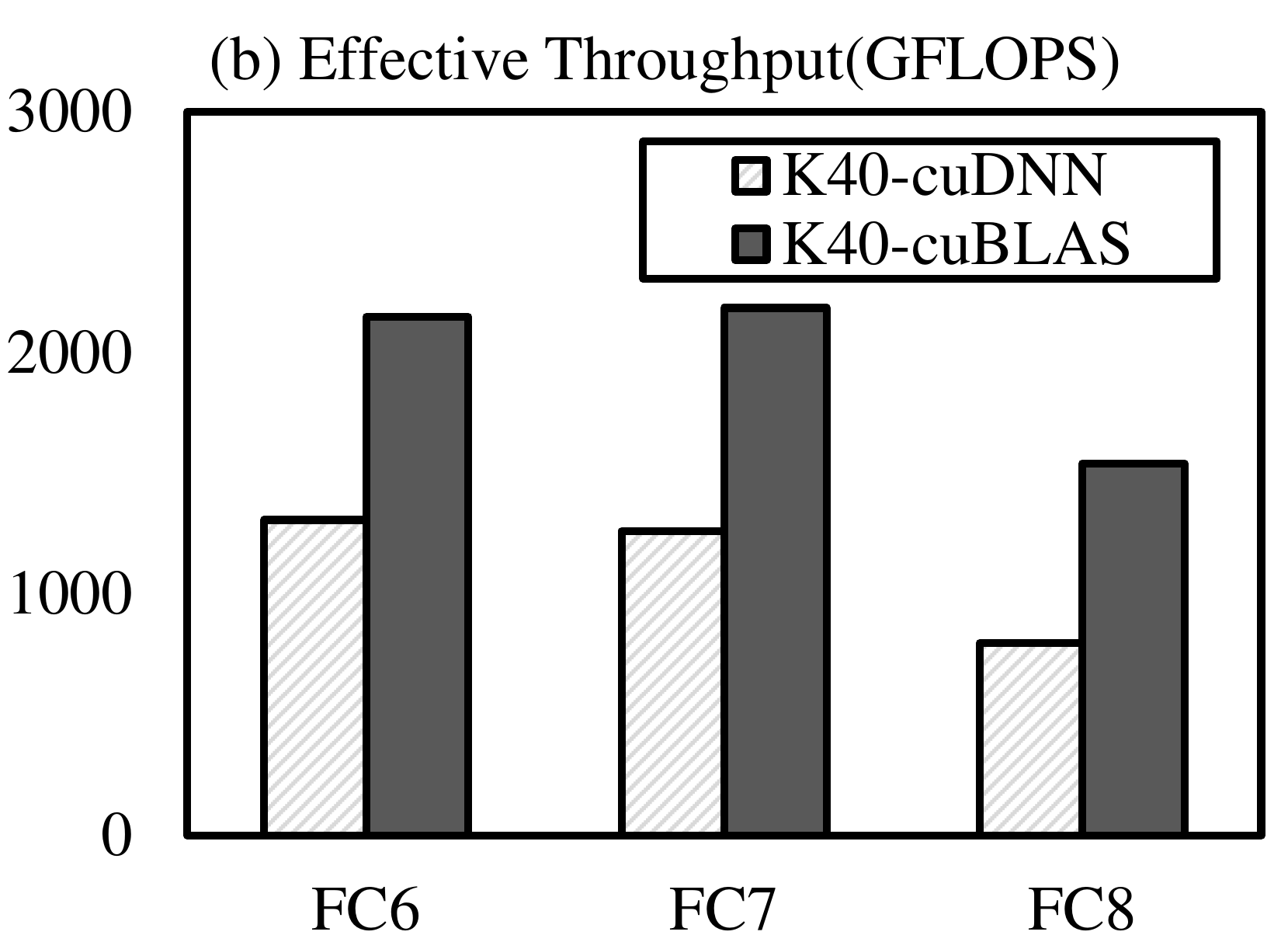}}
			\centerline{}
			\centerline{\includegraphics[width=4.5cm]{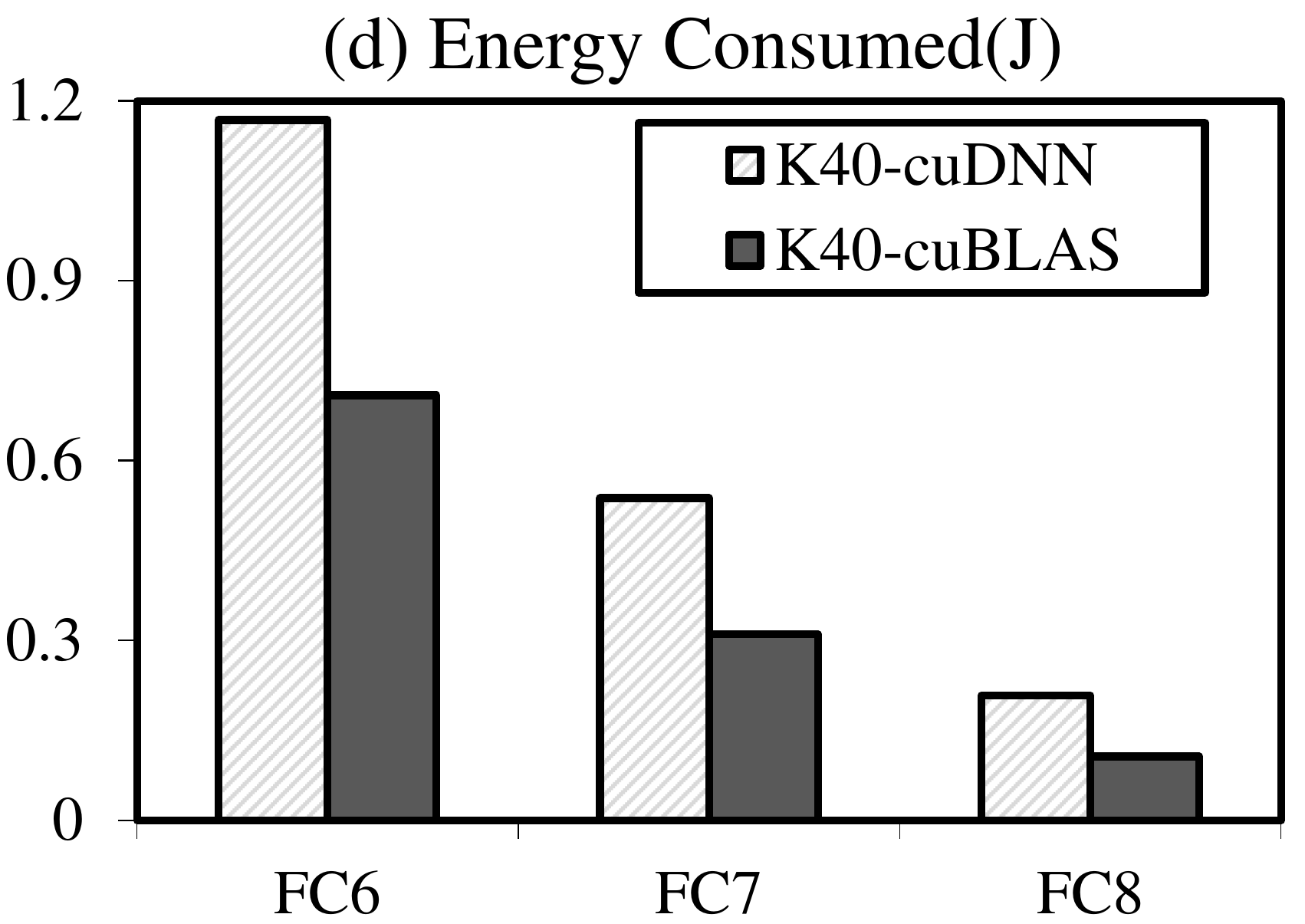}}
			\centerline{}
			\centerline{\includegraphics[width=4.5cm]{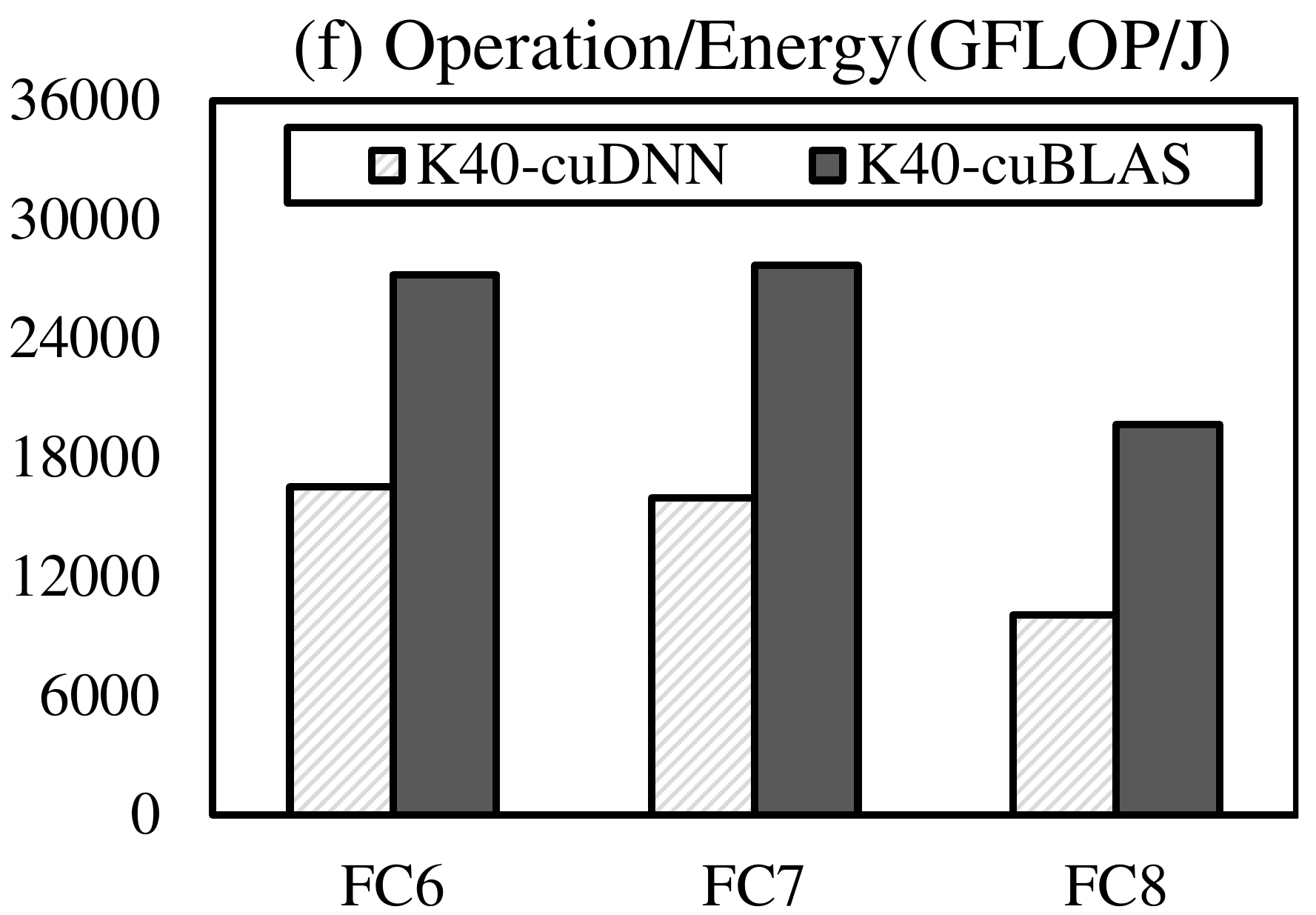}}
		\end{minipage}
		\hfill
		\caption{Forward Comparison between Different GPU Models (cuDNN vs cuBLAS).}
		\label{gpufp}
	\end{figure}

	\begin{table*}[!t]
		\renewcommand{\arraystretch}{1.3}
		\caption{Resource Utilization of the Accelerator on FPGA}
		\label{resource}
		\centering
		\begin{tabular}{|c||c||c||c||c|}
			\hline
			&\textbf{Conv Layer}&    \textbf{LRN}& \textbf{FC} & \textbf{Pooling}\\
			\hline
			ALUTS &    209786    &48327&   112152 & 35247\\         \hline
			Registers    &320,656&   82,469&    197,666 & 54,603\\         \hline
			Logic utilization   &172,006/234,720(73\%)&   51,185/234,720 (22\%)&    99,753/234,720 (42\% )& 40,581/234,720(17\%)\\         \hline
			I/O pin&    279/1,064(26\%)&    279/1,064(26\%)&   279/1,064(26\%) &279 / 1,064(26\%)\\         \hline
			DSP blocks&    162/256(63\%)&   3/256(1\%)&    130/256(51\%) & 0/256(0\%)\\            \hline
			Memory bits& 8,236,663/52,428,800(16\%) & 3,996,240/52,428,800(8\%)& 5,556,688/52,428,800(11\%)&1,419,856/52,428,800(3\%) \\            \hline
			RAM Blocks & 1,428/2,560(56\%) &432/2,560(17\%) & 651/2,560(25\% )&283/2,560(11\% )\\            \hline
			Actual Clock Freq & 171.29MHz & 269.02MHz & 216.16MHz & 304.50MHz \\ \hline
		\end{tabular}
		
	\end{table*}

	\subsection{Comparison between Different GPU Models}
	
	Above results demonstrate that GPU can achieve significantly higher throughput and performance density, especially for the FC layers. In this section, to evaluate the impact of the different GPU library models, we use both cuDNN and cuBLAS library to implement the FC layers in forward computation and back propagation, as illustrated in Table \ref{gpupara}.

\begin{figure}
		\begin{minipage}{0.47\linewidth}
			\centerline{\includegraphics[width=4.5cm]{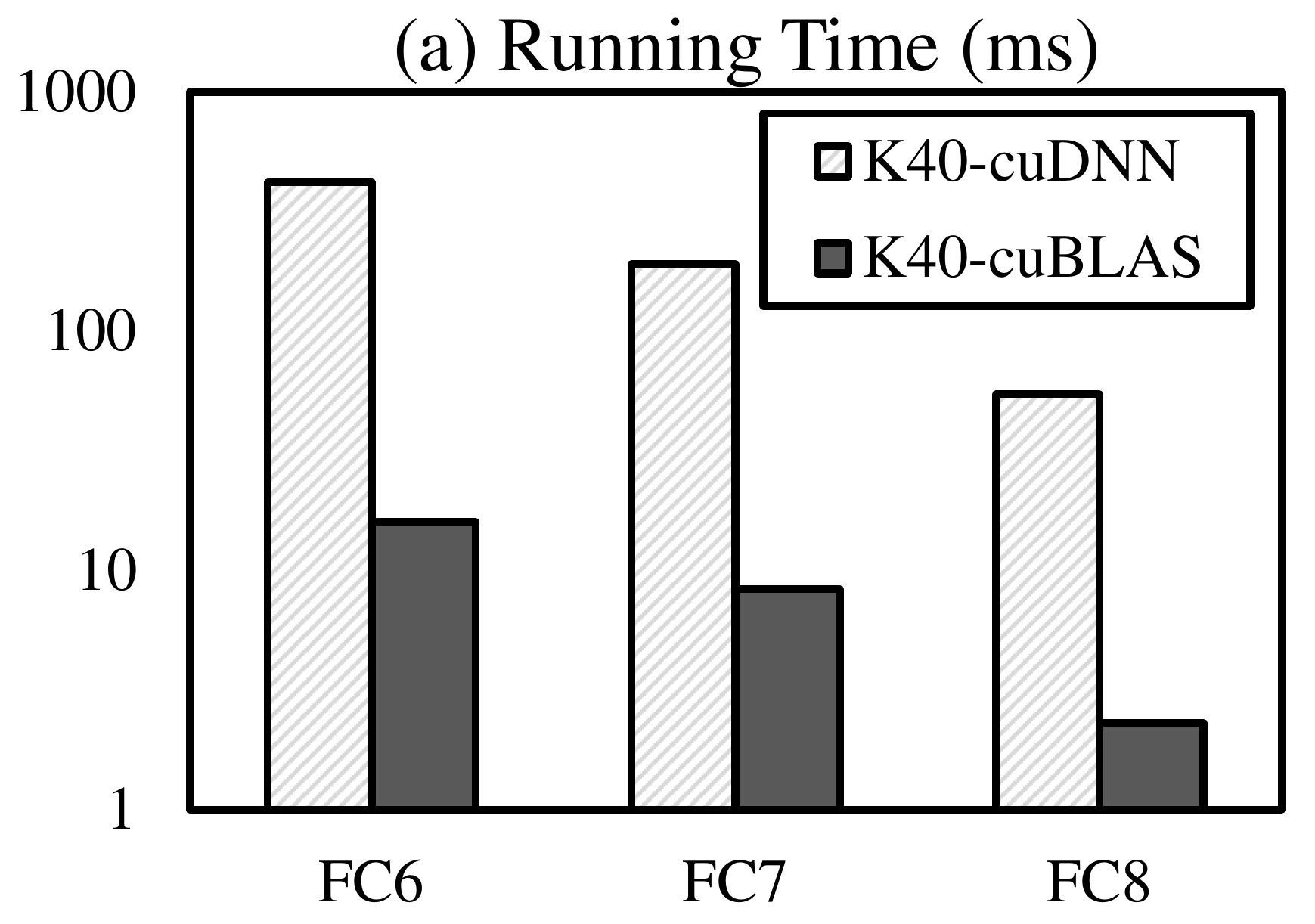}}
			\centerline{}
			\centerline{\includegraphics[width=4.5cm]{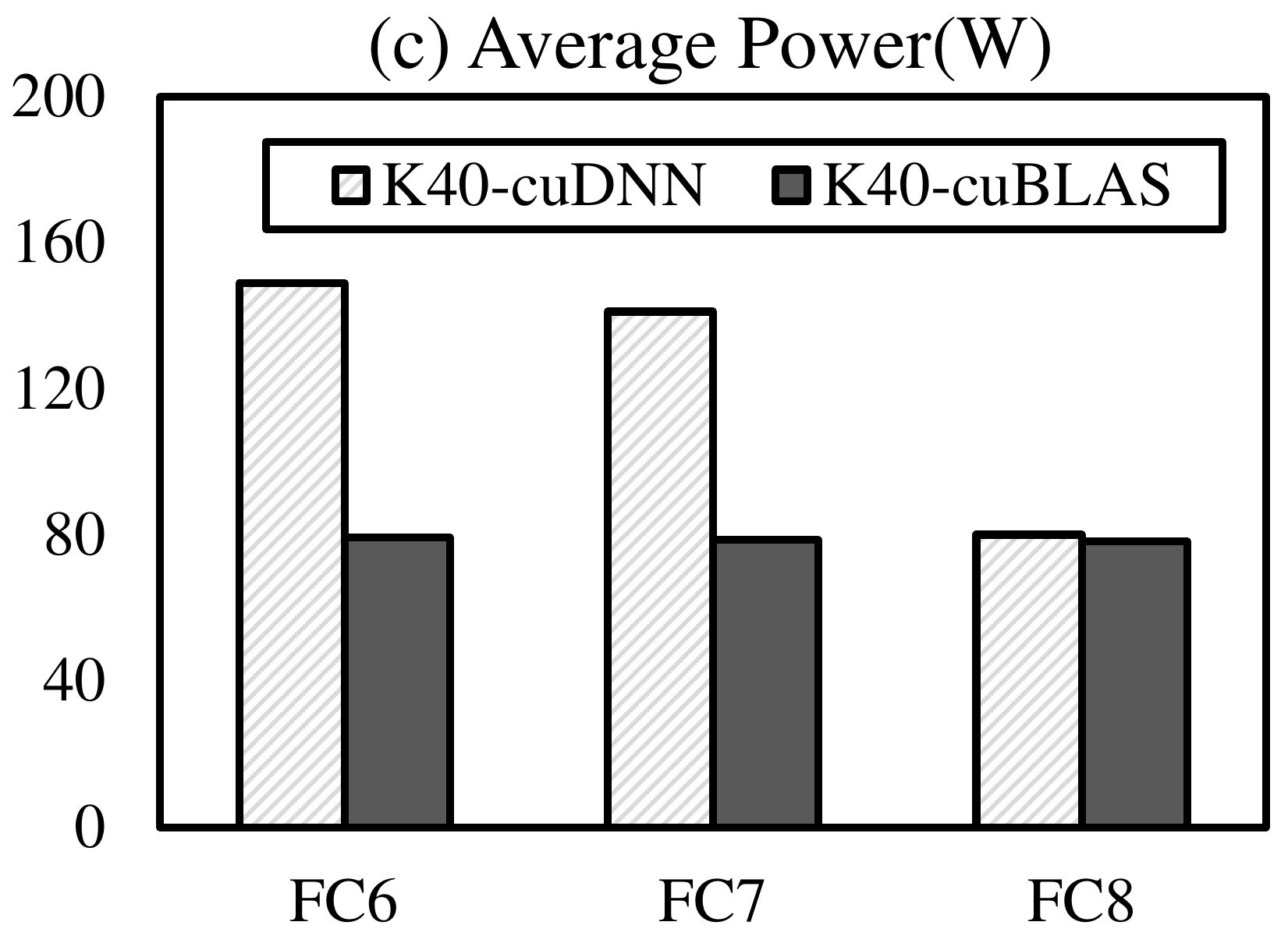}}
			\centerline{}
			\centerline{\includegraphics[width=4.5cm]{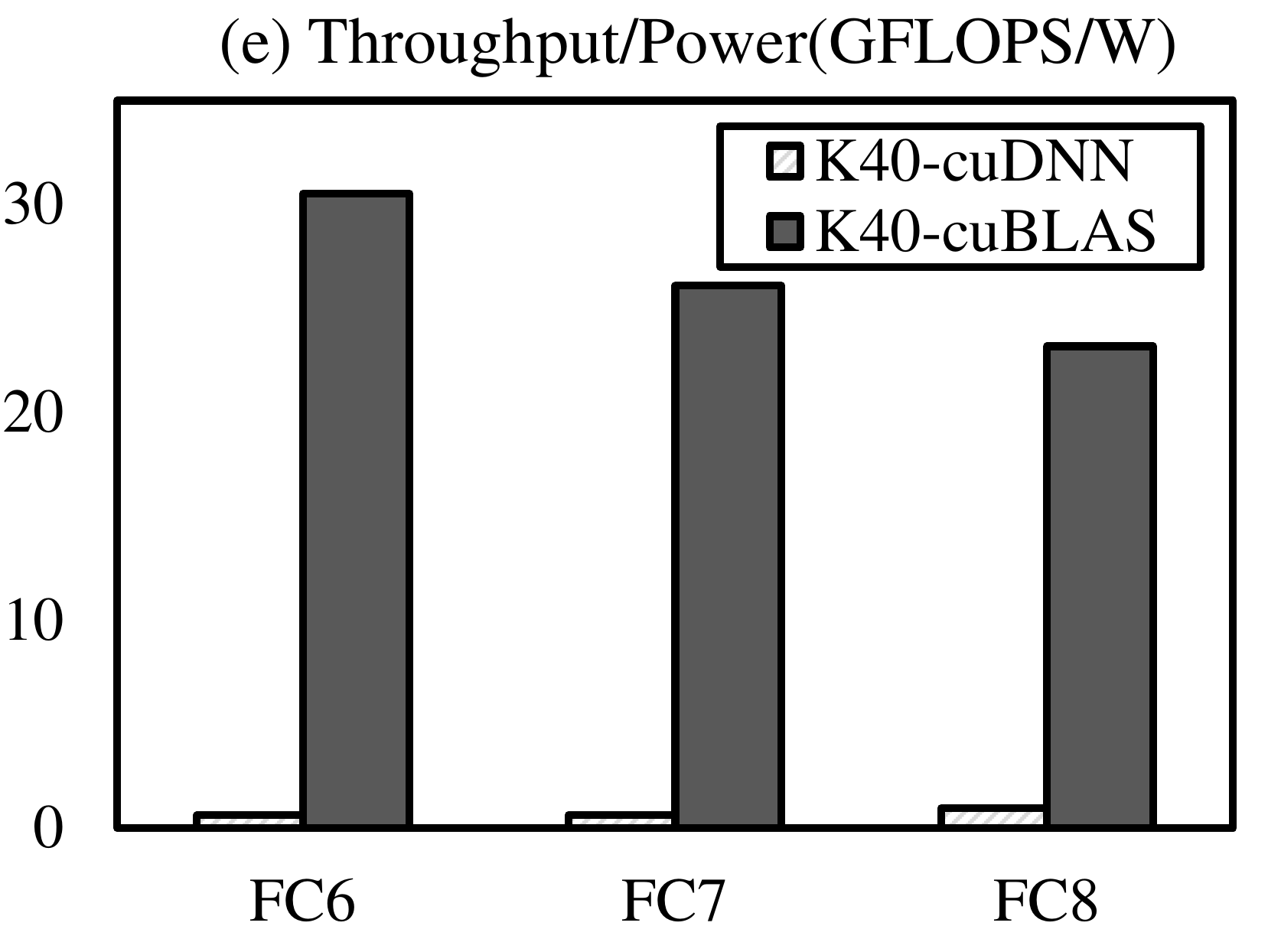}}
		\end{minipage}
		\hfill
		\begin{minipage}{0.47\linewidth}
			\centerline{\includegraphics[width=4.5cm]{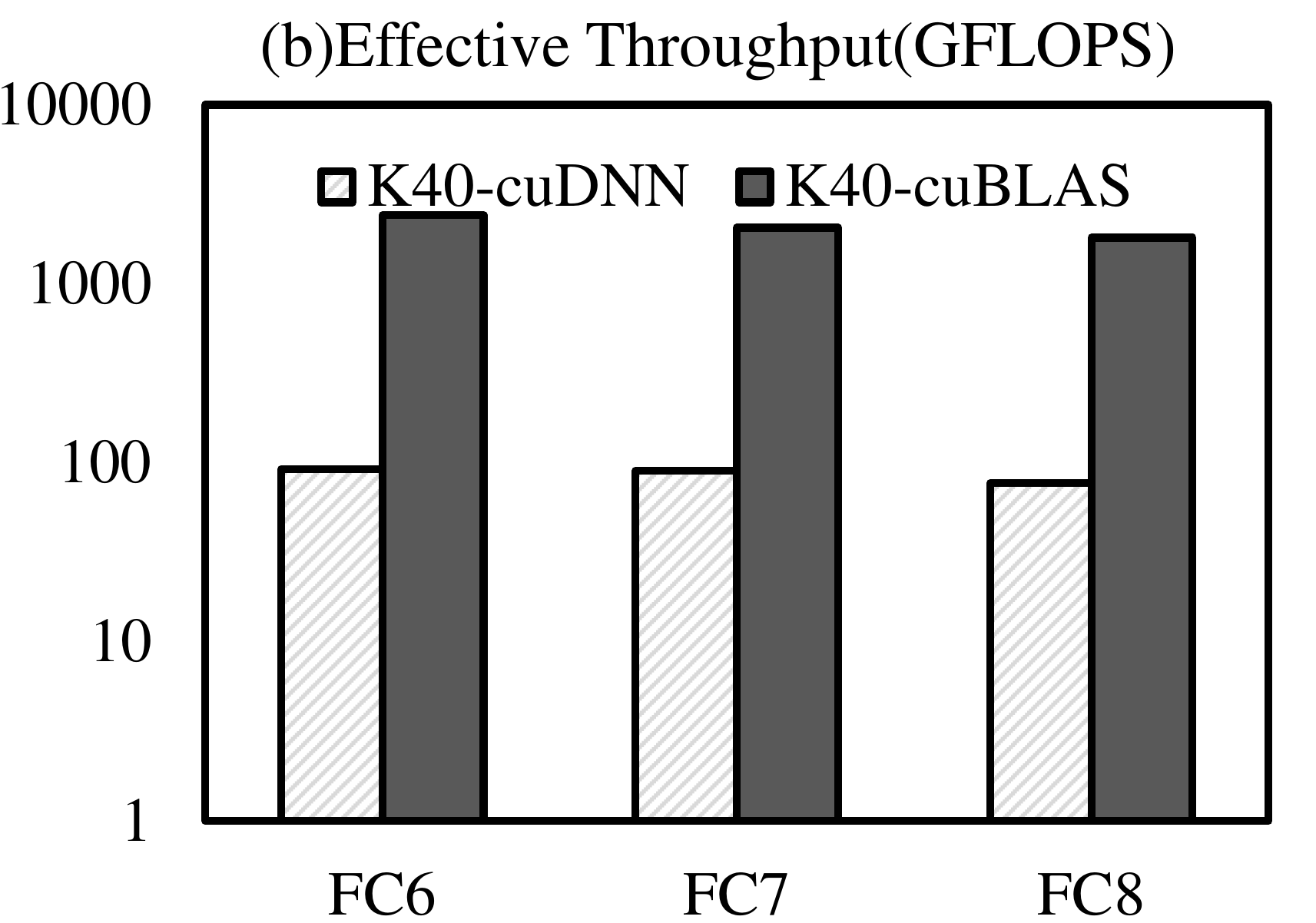}}
			\centerline{}
			\centerline{\includegraphics[width=4.5cm]{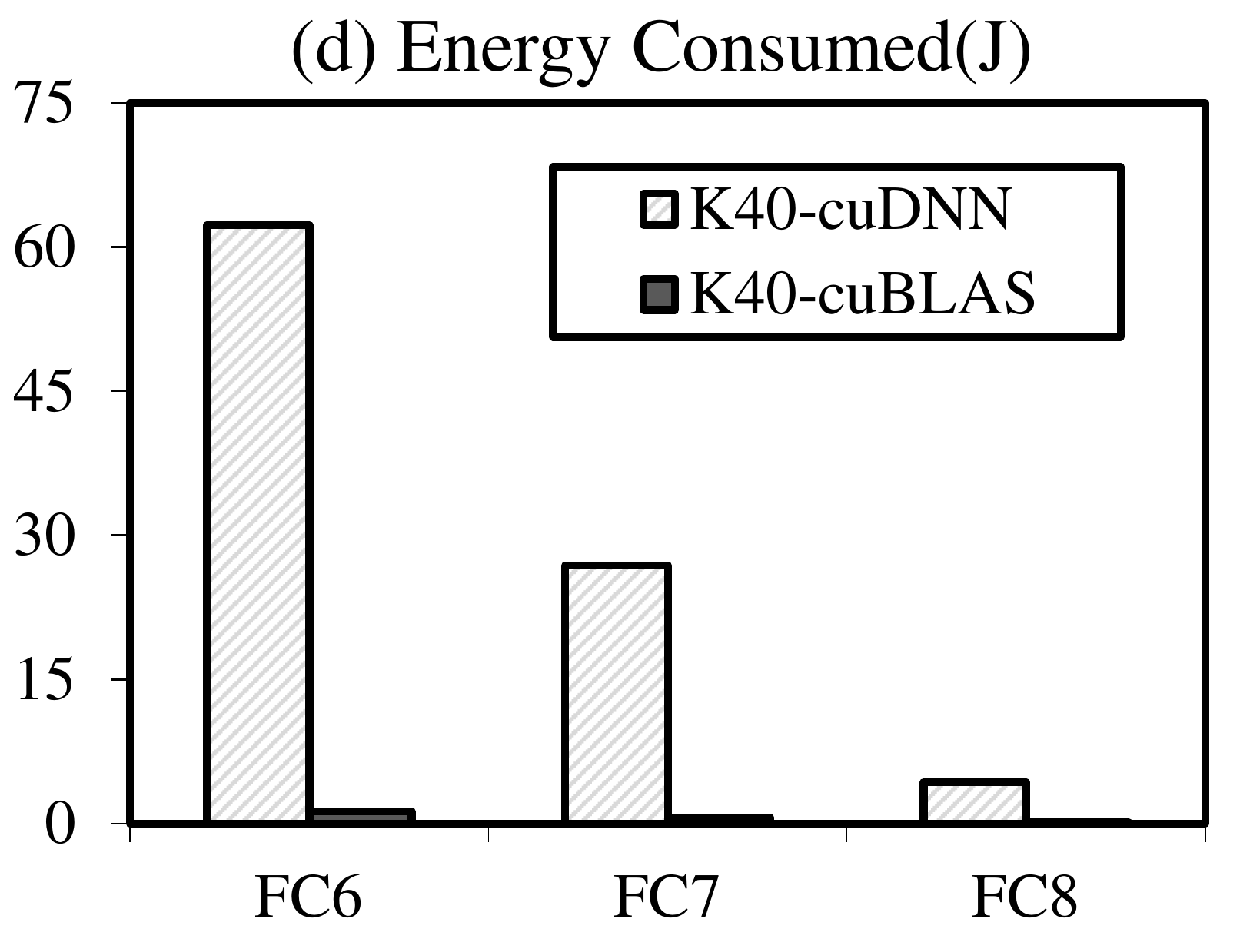}}
			\centerline{}
			\centerline{\includegraphics[width=4.5cm]{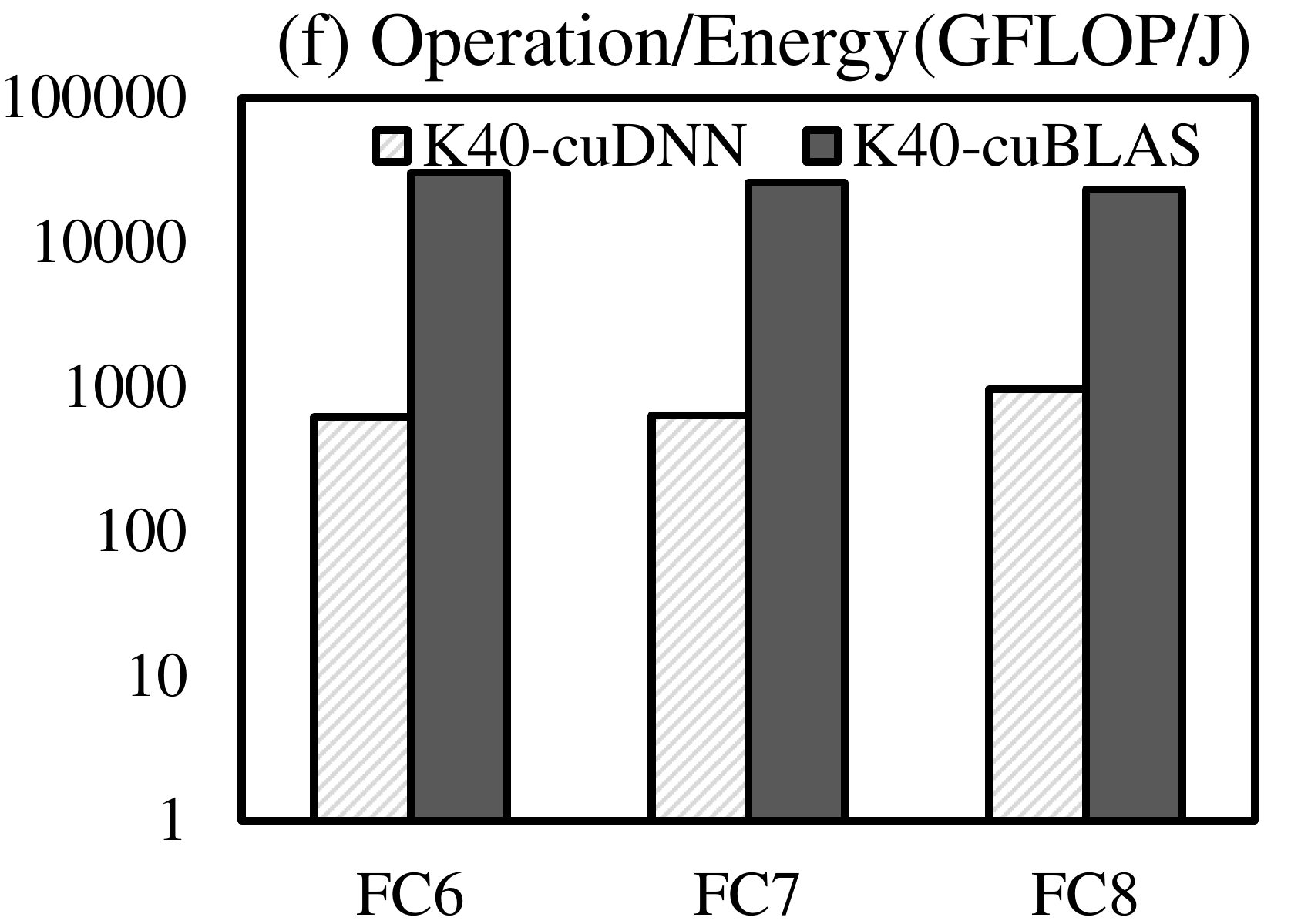}}
		\end{minipage}
		\hfill
		\caption{Backward Comparison between Different GPU Models (cuDNN vs cuBLAS).}
		\label{gpubp}
	\end{figure}

	Fig. \ref{gpufp} and Fig. \ref{gpubp} present the comparison between cuDNN and cuBLAS for forward computation and back propagation, respectively. We have following observations based on the experimental results:
	\begin{itemize}
		\item In general, the cuBLAS library kernels achieve higher speedup (calculated by execution time) than the cuDNN library kernels. In particular, the speedup for cuBLAS against cuDNN is 1.69x in forward computation and 24.89x in BP. In comparison, the throughput for cuBLAS is 1.77x higher than cuDNN in forward computation, but cuDNN achieves 1.57x than cuBLAS in BP calculation.
		\item The cuDNN and cuBLAS libraries have similar power consumptions for forward computation (79.12W and 78.73W on average, respectively), while for the BP, cuBLAS takes significantly more power saving than cuDNN, with the average power 78.77W and 123.40W respectively. Accordingly, the energy consumption of cuBLAS is much lower than the cuDNN, with the average energy 0.70J and 31.19J respectively.
		\item Regarding the performance density, we calculate the Throughput/Power and Operation/Energy accordingly. Results demonstrate cuBLAS substantially outperforms the cuDNN library on performance density metrics.
	\end{itemize}

	\subsection{Resources Usage and Running Frequency}
	
	Table \ref{resource} lists the resources and power consumption for the modules in CNNLab accelerator. In particular, Of these NN layers, convolutional layer takes most significant logic devices as it requires computational power. In particular, the convolutional layer needs 73\% of the hardware logics, 63\% DSP blocks, and 56\% RAM blocks. In comparison, pooling layer only takes 17\% logic resources and 11\% RAM blocks. Regarding the running frequency, the convolutional layer has the lowest frequency at 171.29MHz, while pooling achieves the highest frequency at 304.50MHz accordingly.

	\section{Related Work}

	The neural network model has been an emerging field during the past few years \cite{iclr16}. In this section, we summarize the related acceleration engines, including cloud computing, GPU, and FPGA, respectively.

	\subsection{Cloud based Acceleration}

	Distributed computing platforms have been widely recognized as the scalable, and easy-to-deploy measures \cite{icml12}. Project Adam \cite{osdi14} describes the design and implementation of a distributed system comprised of commodity server machines to train large-scale deep learning models. SINGA \cite{mm15} is a distributed deep learning system for training big models over large datasets. DistBelief \cite{NIPS2012} is a software framework that can utilize computing clusters with a good number of machines to train large models.
	
	\subsection{GPU based Accelerators}
	
	GPU has been widely applied to the acceleration engine for data-intensive applications. For example, Coates et. al \cite{icml2013} present a high-performance computing system with a cluster of GPU servers, using Infiniband interconnects and MPI. NGPU \cite{gpumicro15} brings GPU accelerators together without hindering SIMT execution or adding excessive hardware overhead. Li et al. \cite{ijcnn14} propose an efficient GPU implementation of the large-scale recurrent neural network and demonstrate the power of scaling up the recurrent neural network with GPUs. Vasilache et al. examine the performance profile using fbfft of CNN training on the current generation of GPU \cite{iclr15}. Teng et al. describe an efficient DBN implementation on the GPU, including the pre-training and fine-tuning processes \cite{ijcnn15}. Recently, GeePS is a scalable deep learning architecture on distributed GPUs with specific parameters \cite{geeps}.

	\subsection{FPGA and Hardware based Accelerators}
	
	To overcome the power consumption issue of the GPU and Cloud based frameworks, many developers seek solutions at hardware level \cite{isscc161,isscc162,isscc163}. For the IC based accelerator, Diannao \cite{Diannao} is one of the pioneers works solidifying the neural networks on the hardware circuits. Origami \cite{Origami} present a tape-out accelerator with silicon measurements of power-, area- and I/O efficiency. Meanwhile, FPGA is more flexible due to the integration of the reconfigurable logic devices. Therefore, it can fit changing applications and parameters in neural networks \cite{deepfpga,fpgaConvNet}. For example, Zhang et al. \cite{dnnfpga15} explores the bandwidth for the parameters facing the limitation of an FPGA chip. Suda et al. \cite{caoyufpga16} presents a design space exploration method OpenCL programming model approach, which can explore the trade-offs the parameters in the network topologies.
	
	Besides the ASIC and FPGA-based accelerators, there have been numerous directions using emerging hardware technologies, such as Memristive Boltzmann Machine \cite{mbm}, and Processing-in-Memory techniques \cite{jetc}. Energy efficient inference engine (EIE) uses compression by pruning the redundant connections and having multiple connections share the same weight \cite{isca16}.
	
	\section{Conclusions and Future Work}
	FPGA and GPU have been demonstrated as very powerful and flexible platform for data-intensive neural network processing in machine learning applications. In this paper, we have presented CNNLab, a middleware support for GPU and FPGA-based framework to accelerate the neural network computing models. It can offload the tasks into different accelerators in the guidance of the neural network model and constraints. To achieve the trade-offs between the GPU and FPGA-based platform, we constructed the real prototype using Intel-Altera DE5 FPGA board and Nvidia K40 GPU platform. We measure the execution time, throughput, power consumption, energy cost, and performance density, respectively.
	
	Experimental Results show that the GPU has better speedup (100x) and throughput (100x) against FPGA-based accelerator while FPGA is more power saving (50x) than GPU. More importantly, in our case study, the energy consumption fo GPU and FPGA are similar in convolutional computation, while GPU is more energy efficient in FC layer calculation. Regarding the performance density, both approaches achieve similar Throughput/Power metrics in convolutional layers (10GFLOPS/W for FPGA, and 14GFLOPS/W for GPU), but GPU has higher Operation/Energy than FPGA-based accelerators, especially for FC computation. Regarding the improvement between different GPU CUDA programming models, we also evaluate the metrics for the state-of-the-art cuDNN and cuBLAS, respectively. Results show that cuBLAS is more energy efficient with significantly higher speedup and lower power consumption.
	
	Although the experimental results are inspiring, there are some future promising directions. First, the speedup of the accelerators can be further improved by compressed network models. Second, the hardware accelerator can be assisted with a large scale data processing framework like Spark or TensorFlow platforms.

	\bibliographystyle{ieeetr}
	\bibliography{bare_conf}

\end{document}